\documentclass{IEEEtran}
\usepackage{cite}
\usepackage{amsmath,amssymb,amsfonts}
\pdfoutput=1
\usepackage{algorithmic}
\usepackage{graphicx}
\usepackage{textcomp}
\usepackage{subfigure}
\usepackage{caption}

\def\BibTeX{{\rm B\kern-.05em{\sc i\kern-.025em b}\kern-.08em
    T\kern-.1667em\lower.7ex\hbox{E}\kern-.125emX}}
    
\begin{document}

\title{Adaptive Fusion for RGB-D Salient Object Detection}

\author{Ningning Wang, Xiaojin Gong* \\
College of Information Science \& Electronic Engineering, Zhejiang University, Hangzhou, China.\\
\{wangnn,gongxj\}@zju.edu.cn}
\maketitle

\begin{abstract}
RGB-D salient object detection aims to identify the most visually distinctive objects in a pair of color and depth images. Based upon an observation that most of the salient objects may stand out at least in one modality, this paper proposes an adaptive fusion scheme to fuse saliency predictions generated from two modalities. Specifically, we design a two-streamed convolutional neural network (CNN), each of which extracts features and predicts a saliency map from either RGB or depth modality. Then, a saliency fusion module learns a switch map that is used to adaptively fuse the predicted saliency maps. A loss function composed of saliency supervision, switch map supervision, and edge-preserving constraints is designed to make full supervision, and the entire network is trained in an end-to-end manner. Benefited from the adaptive fusion strategy and the edge-preserving constraint, our approach outperforms state-of-the-art methods on three publicly available datasets.
\end{abstract}

\section{Introduction}
\label{sec:introduction}
Salient object detection aims to automatically identify the most attractive regions in images like human visual systems. It can serve as a useful pre-processing step for various computer vision applications such as image segmentation~\cite{lai2016saliency}, person re-identification~\cite{bi2014person}, object localization~\cite{cinbis2014multi} and tracking~\cite{wu2017vision}, and therefore has received considerable attention. Although great progress has been made in this field, most works~\cite{judd2009learning, liu2011learning, cheng2015global, li2015visual, wang2015deep, li2016deep, hou2017deeply} focus merely on color images. When objects share similar appearances with their surroundings or present with complex background, the algorithms based on color images often fail to distinguish them as salient objects. 

The above-mentioned challenges can be overcomed to a large extent if depth information is available. In recent years, robust ranging sensors such as stereo cameras, RGB-D cameras, and lidars make it easy to collect paired color and depth images. RGB-D saliency detection has consequently been attracting more and more research interest. Published literatures made efforts on modeling depth-induced saliency detection~\cite{ju2014depth, feng2016local, ren2015exploiting} and fusing multi-modalities~\cite{peng2014rgbd, qu2017rgbd, han2017cnns, chen2019multi, chen2018progressively}. However, existing works performed fusion via either directly concatenating color and depth features, or element-wise multiplication/addition of predictions generated by the two modalities. Such fusion strategies are inadequate to combine complementary information from two modalities, leaving a room for performance improvement.

\begin{figure*}[htbp] 
\centering
\subfigure[RGB]{
\centering
\begin{minipage}{0.13\linewidth}
\centering
\includegraphics[width=1\columnwidth]{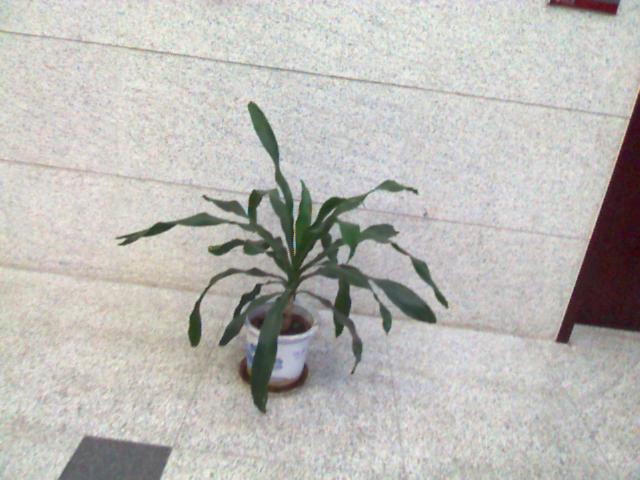} \\\vspace{1.5pt}
\includegraphics[width=1\columnwidth]{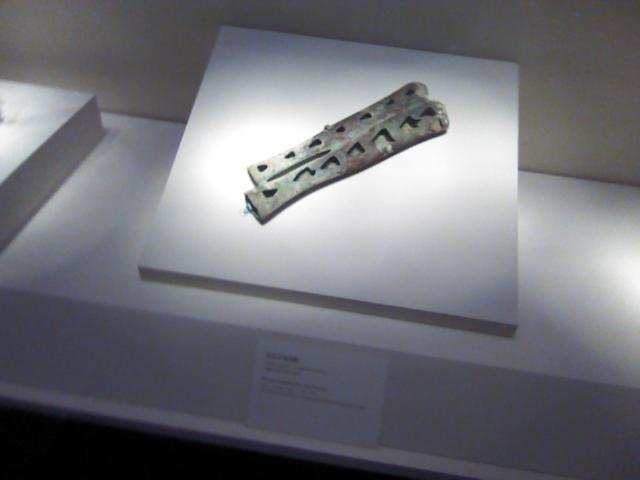} \\\vspace{1.5pt}
\includegraphics[width=1\columnwidth]{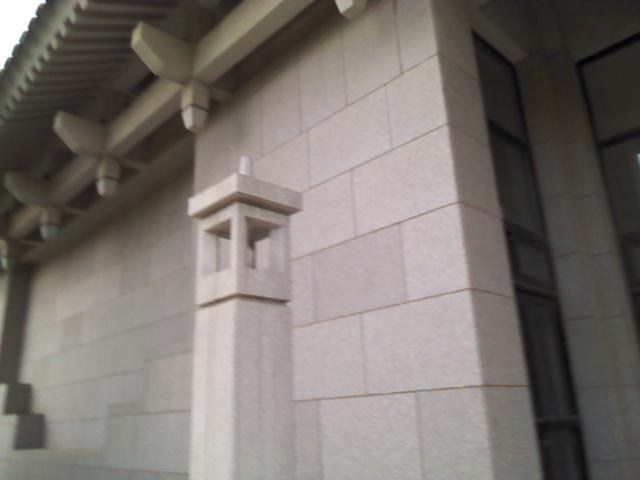} \\\vspace{1.5pt}
\includegraphics[width=1\columnwidth]{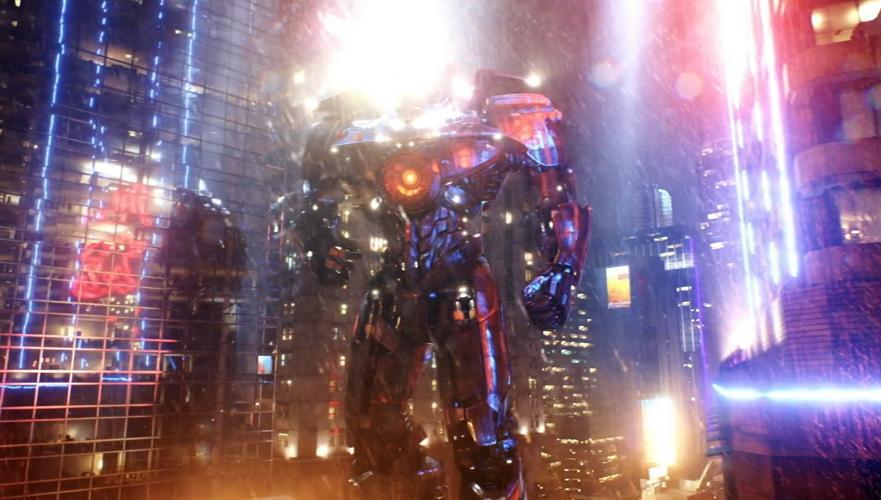} \\\vspace{1.5pt}
\end{minipage} 
}
\hspace{-12pt}
\subfigure[Depth]{
\centering
\begin{minipage}{0.13\linewidth}
\centering
\includegraphics[width=1\columnwidth]{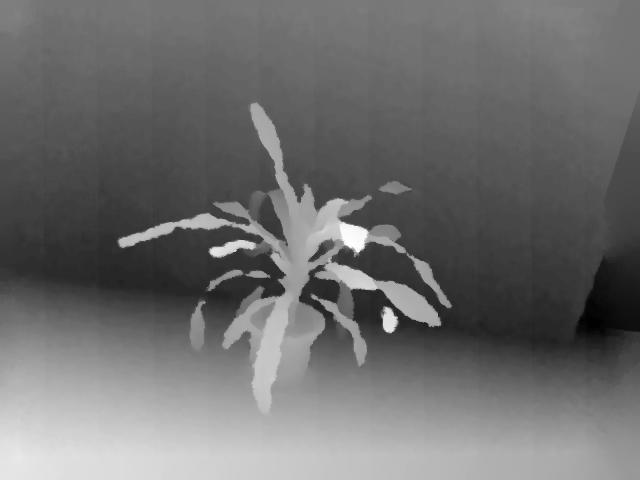} \\\vspace{1.5pt}
\includegraphics[width=1\columnwidth]{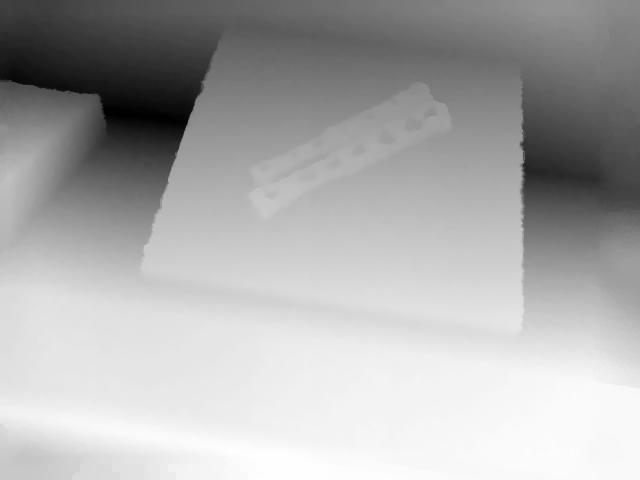} \\\vspace{1.5pt}
\includegraphics[width=1\columnwidth]{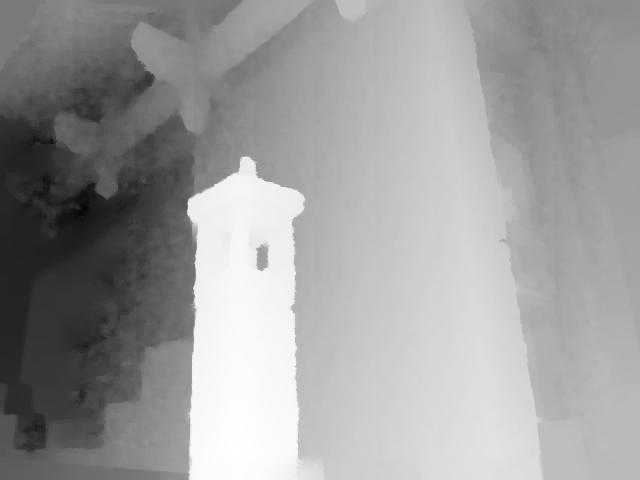} \\\vspace{1.5pt}
\includegraphics[width=1\columnwidth]{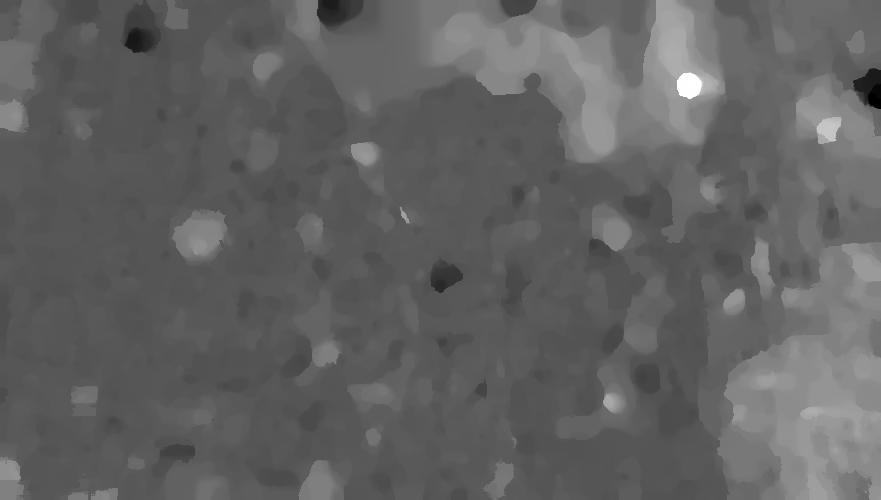} \\\vspace{1.5pt}
\end{minipage} 
}
\hspace{-12pt}
\subfigure[Ground truth]{
\centering
\begin{minipage}{0.13\linewidth}
\centering
\includegraphics[width=1\columnwidth]{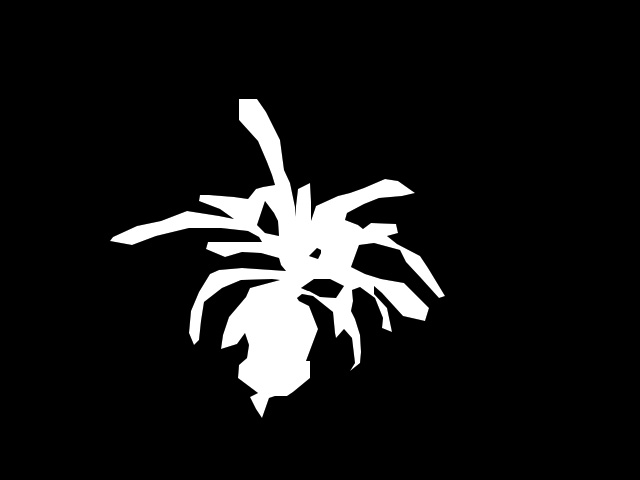} \\\vspace{1.5pt}
\includegraphics[width=1\columnwidth]{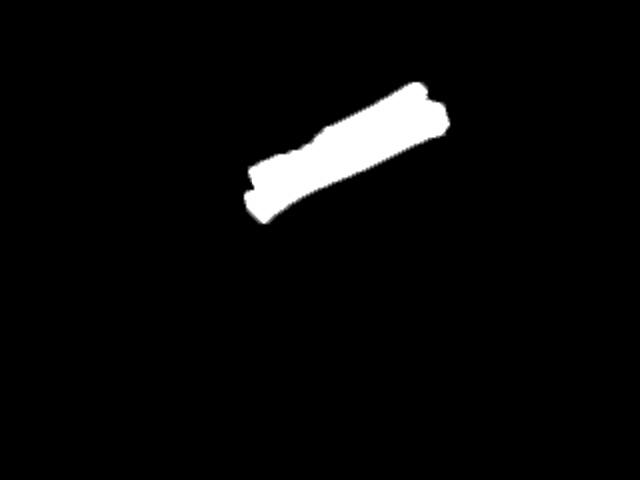} \\\vspace{1.5pt}
\includegraphics[width=1\columnwidth]{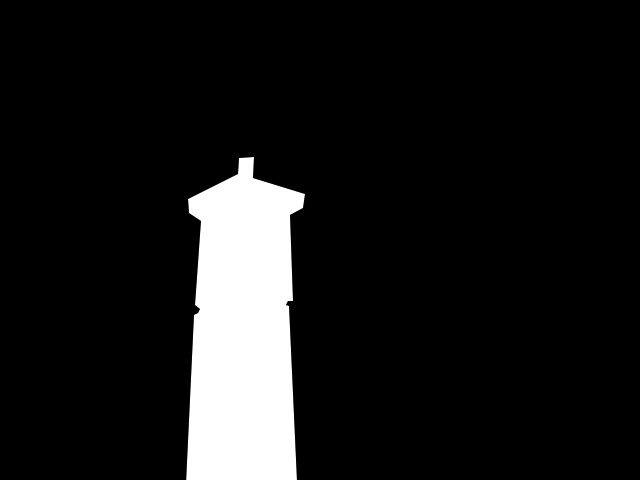} \\\vspace{1.5pt}
\includegraphics[width=1\columnwidth]{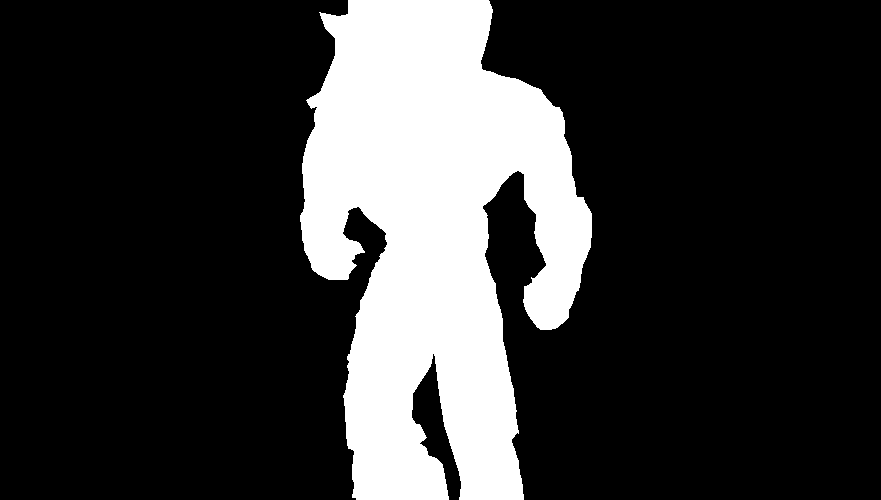} \\\vspace{1.5pt}
\end{minipage} 
}
\hspace{-12pt}
\subfigure[$\mathbf{S}^{rgb}$]{
\centering
\begin{minipage}{0.13\linewidth}
\centering
\includegraphics[width=1\columnwidth]{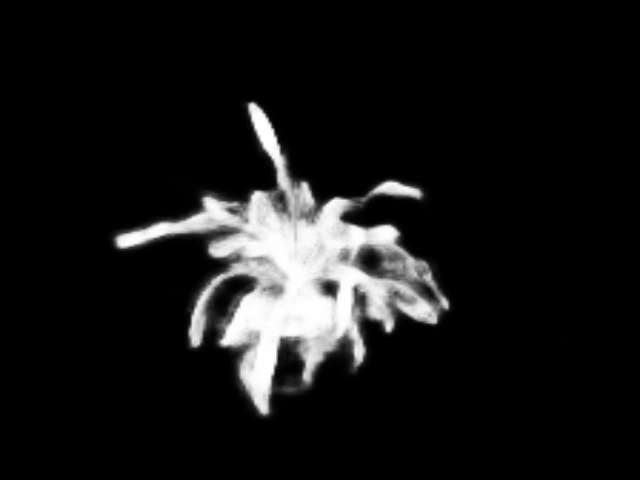} \\\vspace{1.5pt}
\includegraphics[width=1\columnwidth]{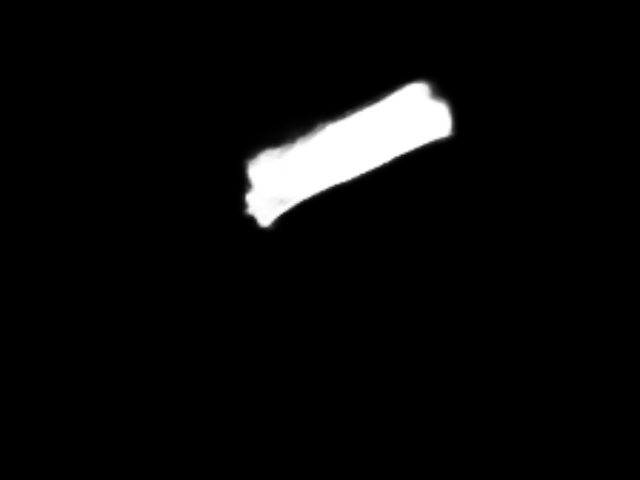} \\\vspace{1.5pt}
\includegraphics[width=1\columnwidth]{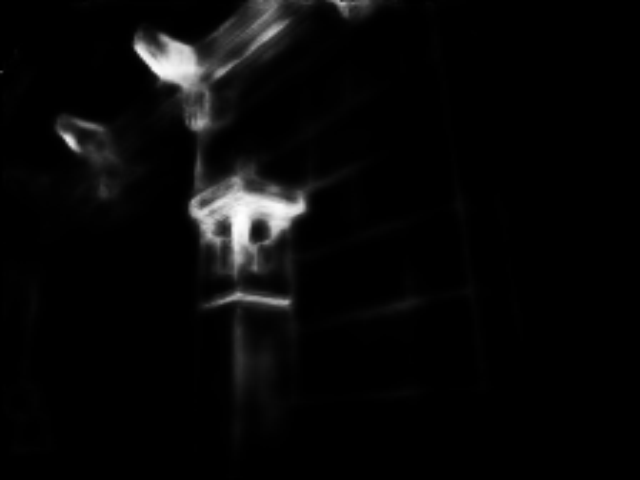} \\\vspace{1.5pt}
\includegraphics[width=1\columnwidth]{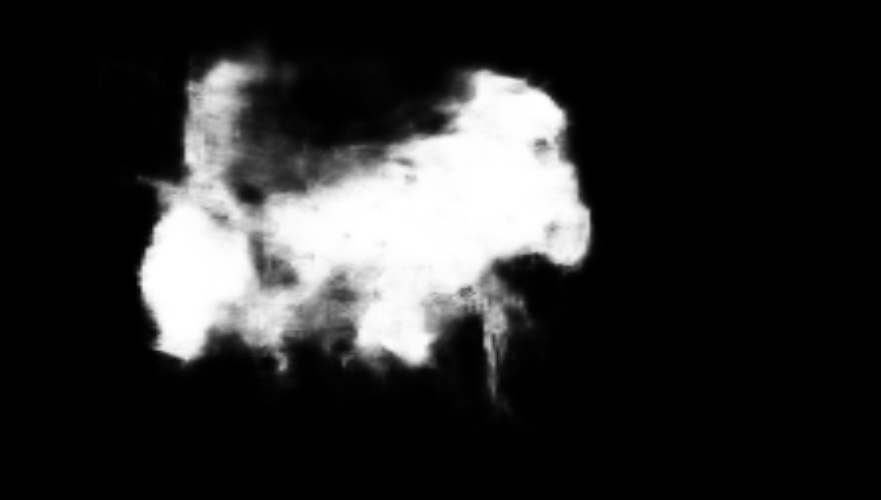} \\\vspace{1.5pt}
\end{minipage} 
}
\hspace{-12pt}
\subfigure[$\mathbf{S}^d$]{
\centering
\begin{minipage}{0.13\linewidth}
\centering
\includegraphics[width=1\columnwidth]{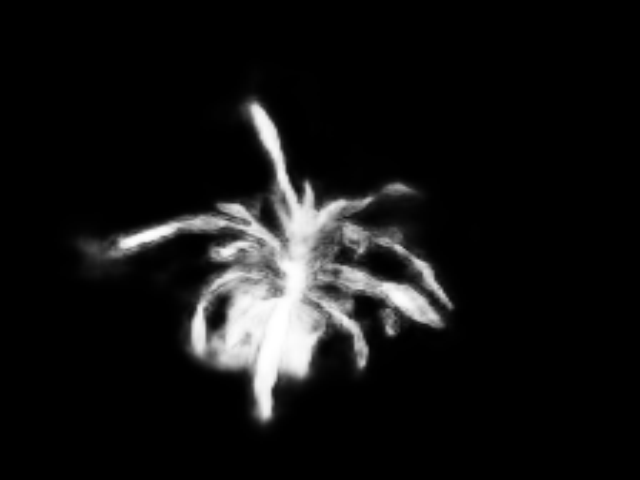} \\\vspace{1.5pt}
\includegraphics[width=1\columnwidth]{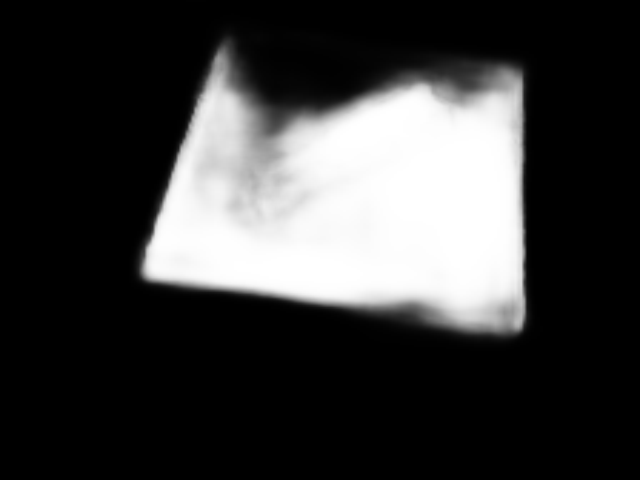} \\\vspace{1.5pt}
\includegraphics[width=1\columnwidth]{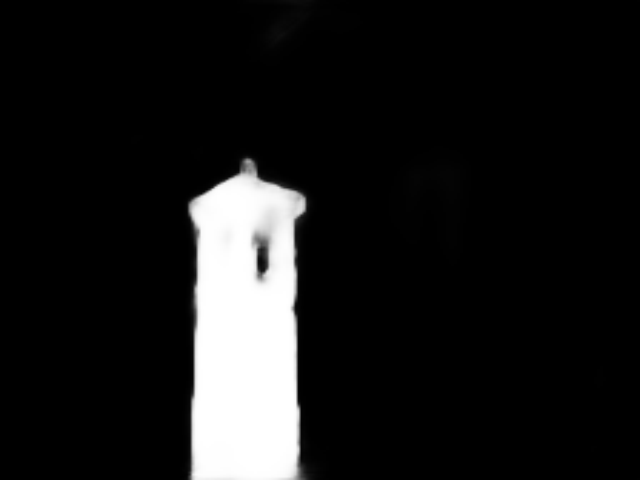} \\\vspace{1.5pt}
\includegraphics[width=1\columnwidth]{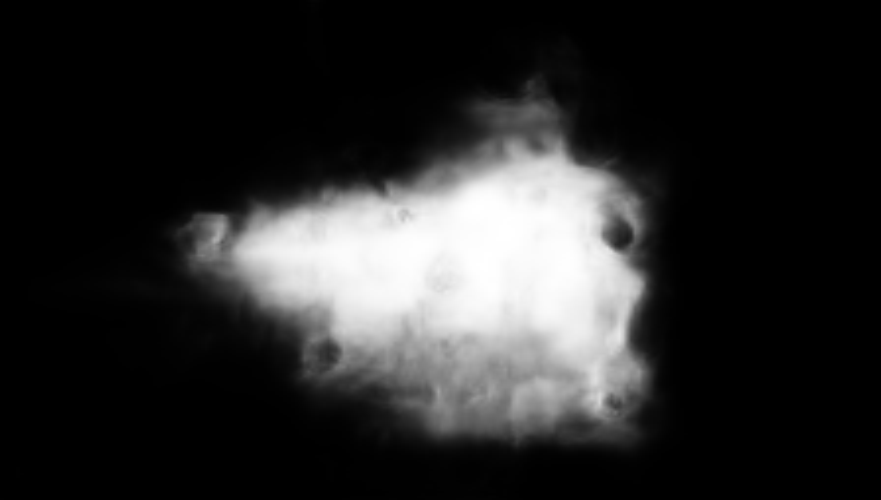} \\\vspace{1.5pt}
\end{minipage} 
}
\hspace{-12pt}
\subfigure[$\mathbf{S}^{fused}$]{
\centering
\begin{minipage}{0.13\linewidth}
\centering
\includegraphics[width=1\columnwidth]{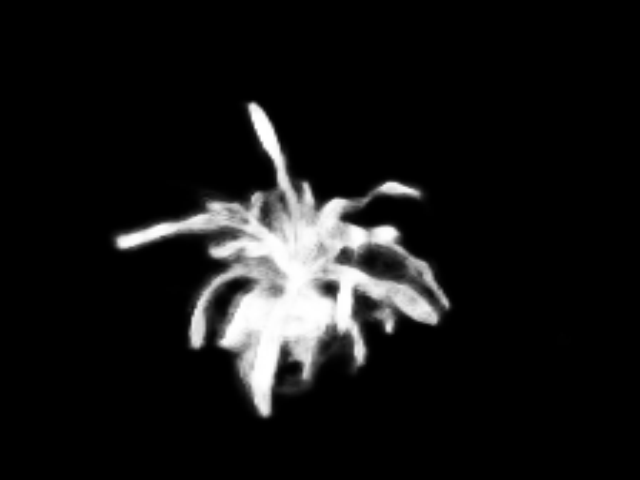} \\\vspace{1.5pt}
\includegraphics[width=1\columnwidth]{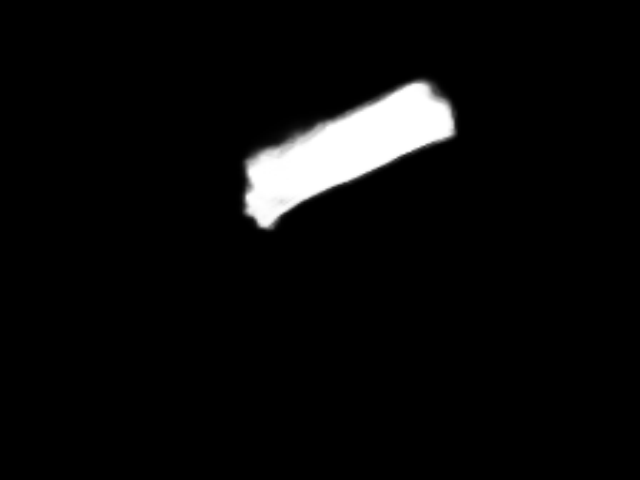} \\\vspace{1.5pt}
\includegraphics[width=1\columnwidth]{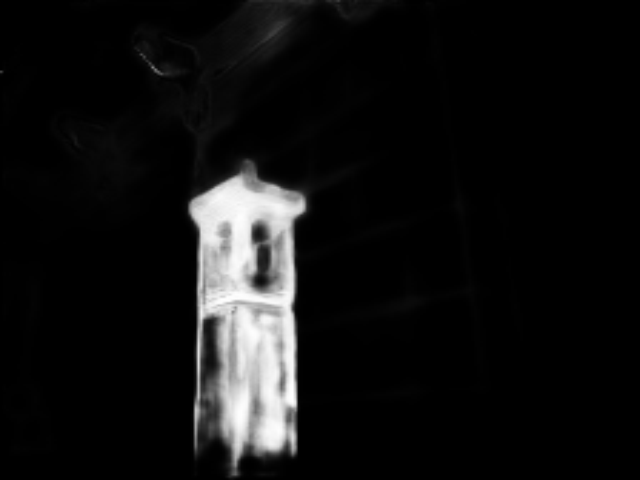} \\\vspace{1.5pt}
\includegraphics[width=1\columnwidth]{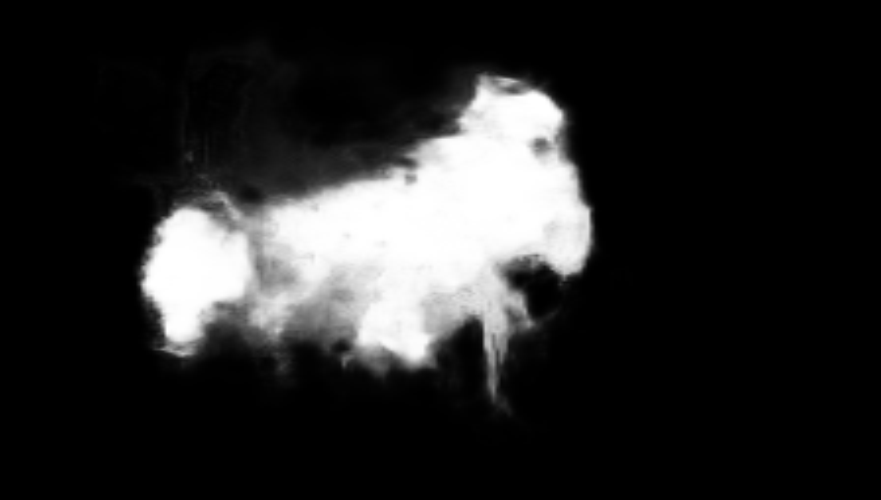} \\\vspace{1.5pt}
\end{minipage} 
}
\hspace{-12pt}
\subfigure[Switch map]{
\centering
\begin{minipage}{0.13\linewidth}
\centering
\includegraphics[width=1\columnwidth]{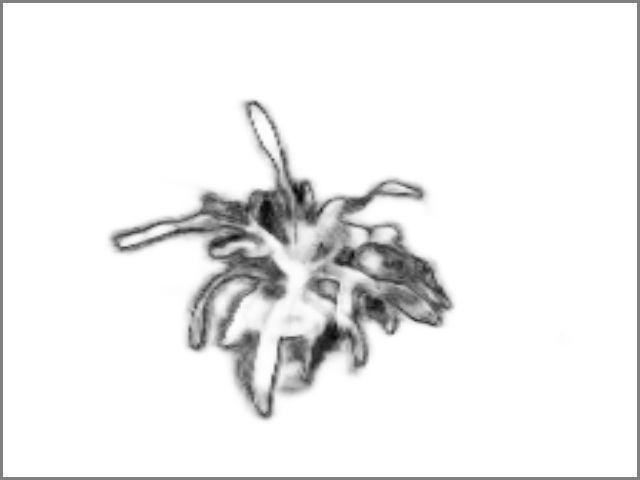} \\\vspace{1.5pt}
\includegraphics[width=1\columnwidth]{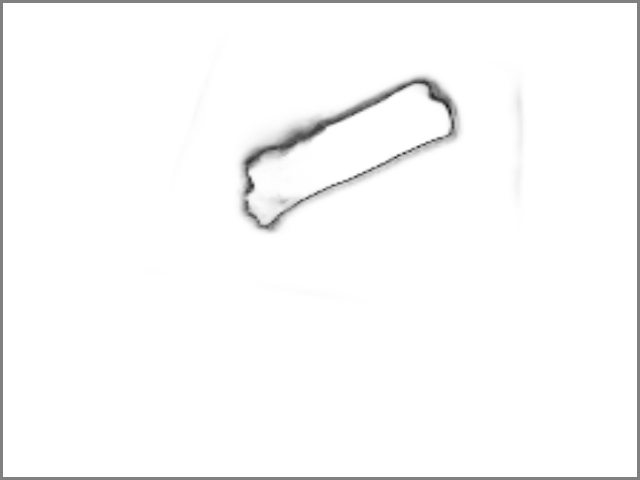} \\\vspace{1.5pt}
\includegraphics[width=1\columnwidth]{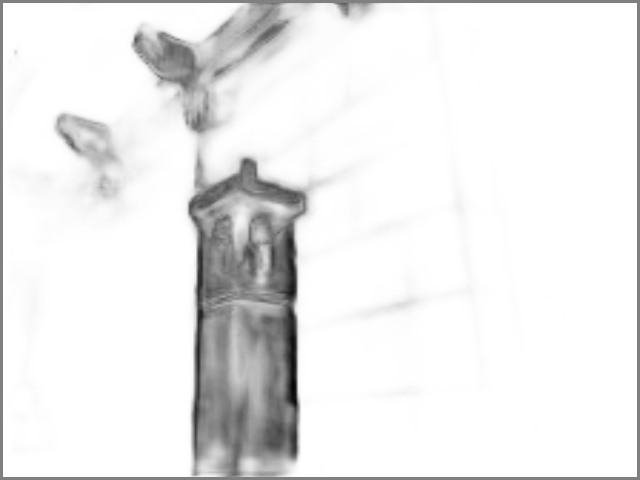} \\\vspace{1.5pt}
\includegraphics[width=1\columnwidth]{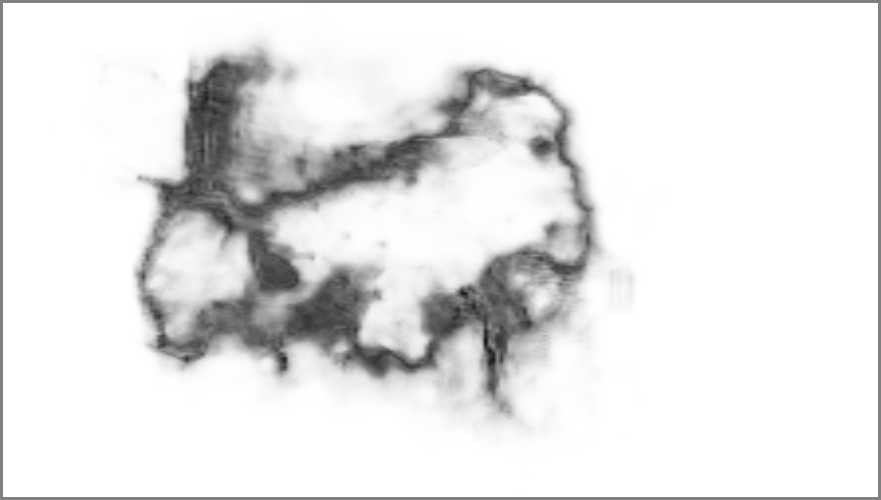} \\\vspace{1.5pt}
\end{minipage} 
}
\caption{Typical scenarios in RGB-D saliency object detection. Here, $\mathbf{S}^{rgb}$ denotes the result obtained by our RGB saliency prediction stream, $\mathbf{S}^{d}$ is the result from our depth saliency prediction stream, and $\mathbf{S}^{fused}$ is the final saliency detection result. Switch map is the map learned in our network for adaptive fusion.}     
\label{fig:SW}     
\end{figure*}

When observing objects in paired color and depth images, we roughly classify scenes into four categories: 1) Objects have distinguishable appearances in both modalities; 2) Objects have close depth values but distinguishable color appearances with backgrounds; 3) Objects share similar color appearances with backgrounds but have different depth values; and 4) Scenes are cluttered in both color and depth images, as shown in Figure~\ref{fig:SW}. For the first three scenarios, salient objects can be correctly detected at least in one modality when using state-of-the-art single-model based saliency detection methods. It implies that good results can be obtained for these scenarios if an algorithm could adaptively choose the predictions from one or the other modality.

Our work is inspired by the above-mentioned observation. In order to make fusion adaptive, We propose an end-to-end framework that consists of a two-streamed convolutional neural network (CNN) and a saliency fusion module to predict and fuse saliency predictions. Our main contributions lie in the following aspects:
\begin{itemize}
	\item We design a two-streamed CNN to predict a saliency map from each modality separately. Each unimodal saliency prediction stream adopts a multi-scale feature aggregation strategy to make feature extraction and saliency prediction effective, while keeping the architecture simple.
	\item We propose a saliency fusion module that learns a switch map to adaptively fuse the predicted saliency maps. A pseudo ground truth switch map is constructed to supervise the learning so that the learned switch map can predict the weights for fusing RGB and depth saliency maps.
	\item The proposed approach is validated on three publicly available datasets, including NJUD~\cite{ju2014depth}, NLPR~\cite{peng2014rgbd}, and STEREO~\cite{niu2012leveraging}. Experimental results show that our approach consistently outperforms state-of-the-art methods on all datasets.
	\item To make our work reproducible, we release our source code at \underline{https://github.com/Lucia-Ningning/Adaptive\_}\break\underline{Fusion\_RGBD\_Saliency\_Detection}.
\end{itemize}

%
\section{Related Work}
\subsection{RGB Saliency Detection}
A great number of RGB salient object detection methods have been developed over the past decades. Traditional methods mainly rely on hand-crafted features and commit to mining effective priors such as center prior~\cite{judd2009learning}, contrast prior~\cite{liu2011learning,cheng2015global}, boundary and connectivity prior~\cite{zhu2014saliency}. Owing to the deep learning revolution, CNN-based approaches have refreshed the previous records in recent years. Multi-scale features are first extracted by multiple CNNs and concatenated together, and then they are fed into a shallow network to predict saliency~\cite{li2015visual}. Whereafter, two-branched networks~\cite{li2016deep}~\cite{wang2015deep} were designed to capture global and local context. In more recent years, deep hierarchical saliency networks (DHSNet)~\cite{liu2016dhsnet}, short connections~\cite{hou2017deeply}, and even more complicated structures such as Amulet~\cite{zhang2017amulet} and agile Amulet~\cite{zhang2018rgb} were developed to aggregate multi-scale features progressively and predict saliency within end-to-end frameworks. We adopt the progressive multi-scale feature aggregation strategy in our unimodal saliency prediction stream, but we keep the network as simple as possible.

\subsection{RGB-D Saliency Detection}
There are two major concerns existing in RGB-D saliency detection: 1) how to model the depth-induced saliency; and 2) how to fuse RGB and depth modalities for achieving better performance. 

Regarding to the first problem, different features such as anisotropic center-surround difference~\cite{ju2014depth} and local background enclosure (LBE)~\cite{feng2016local} were designed to evaluate saliency on depth maps. Global priors, including the normalized depth prior and the global-context surface orientation prior~\cite{ren2015exploiting}, were exploited as well. Although these features and priors are particularly effective for depth saliency detection, their performances are limited due to hand-crafted designs and multi-stage models.  

\begin{figure*}[tb] 
\centering 
\includegraphics[width=1.8\columnwidth]{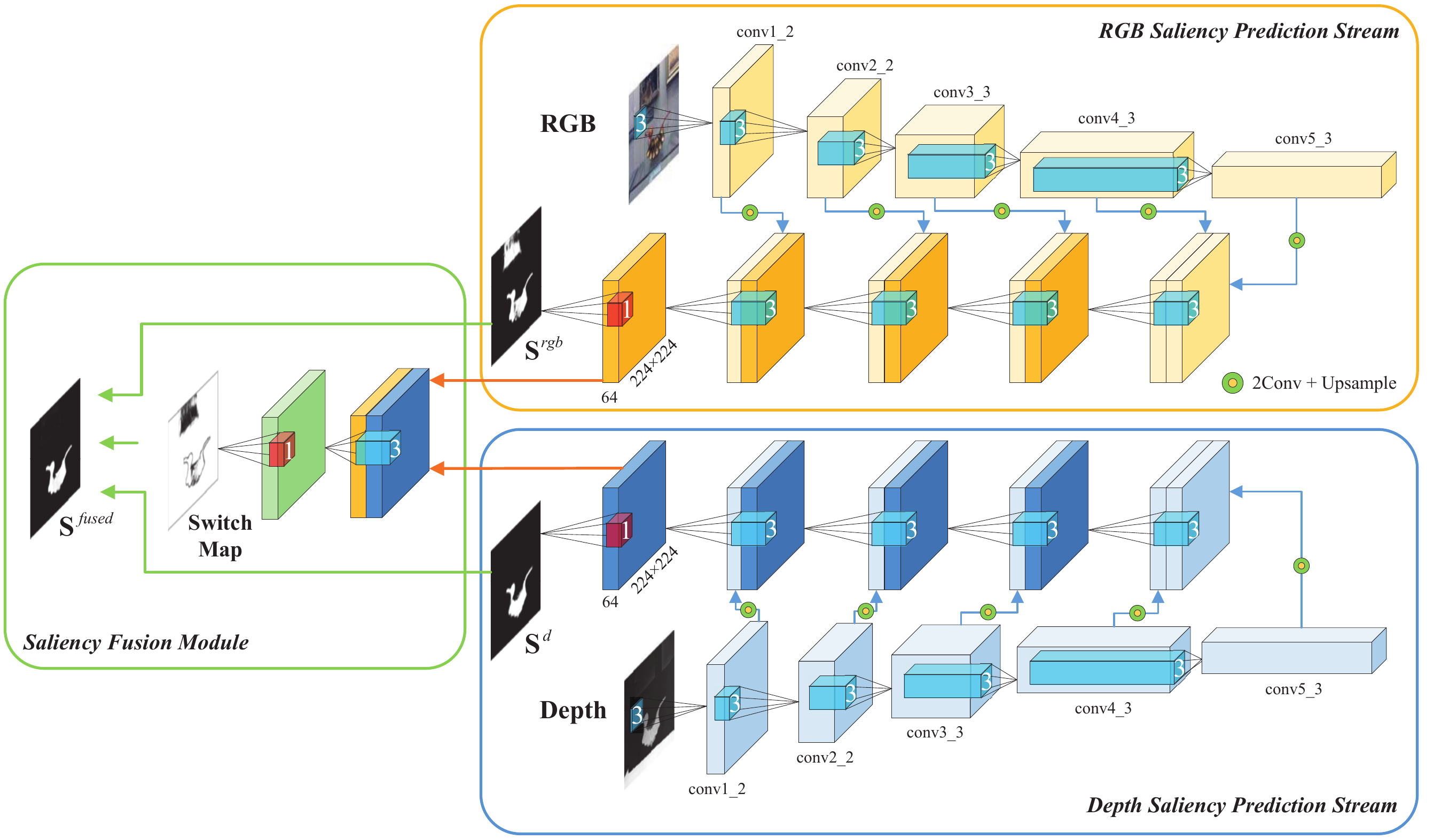}    
\caption{The overview of our framework for RGB-D salient object detection.\label{fig:architecture}} 
\end{figure*}

For the second problem, existing approaches perform multi-modal fusion roughly at input, feature, or decision levels. For instance, Peng et al.~\cite{peng2014rgbd} directly concatenated RGB and depth values and fed the 4-channel data into a multi-stage model to produce saliency maps. Qu et al.~\cite{qu2017rgbd} extracted hand-crafted features from RGB-D superpixels and input them into a shallow CNN for feature combination and saliency regression. Han et al.~\cite{han2017cnns} designed a two-streamed CNN to extract RGB and depth features separately and then fuse them with a joint representation layer. Fusion in these methods is conducted with a single path. In order to enable sufficient fusion, a multi-scale multi-path fusion network~\cite{chen2019multi} and a progressively complementarity-aware fusion network~\cite{chen2018progressively} were developed in more recent years. Although complicated architectures were designed, these methods perform fusion mainly rely on feature concatenation and element-wise addition/multiplication of prediction results. In contrast, we design an adaptive fusion scheme to fuse prediction results from RGB and depth modalities and achieve better detection performance.

\section{The Proposed Method}
When a pair of RGB and depth images are given, we feed them into a two-streamed network for saliency detection. In each stream, features at different scales are progressively aggregated and a saliency map, $\mathbf{S}^{rgb}$ or $\mathbf{S}^d$, is predicted based upon unimodal information. In addition, the last layer of RGB and depth features are concatenated to generate a switch map $\mathbf{SW}$. The switch map further explicitly guides the fusion of $\mathbf{S}^{rgb}$ and $\mathbf{S}^d$ to produce the final saliency map $\mathbf{S}^{fused}$. During training, all of the predicted saliency maps are supervised under the ground truth $\mathbf{Y}$ and the switch map is supervised with a pseudo ground truth constructed from $\mathbf{Y}$ and $\mathbf{S}^{rgb}$. Figure~\ref{fig:architecture} illustrates an overview of the proposed framework.

\subsection{Unimodal Saliency Prediction Stream}
A unimodal saliency prediction stream aims to predict a saliency map based upon a single modal information. Therefore, the design can be benefited from state-of-the-art RGB saliency detection methods. Our design adopts a multi-scale feature based saliency detection framework~\cite{liu2016dhsnet, hou2017deeply, zhang2017amulet, zhang2018rgb}, using an effective feature fusion strategy that progressively aggregates multi-scale features. In contrast to these methods, we keep our network structure as simple as possible.

Specifically, each stream is built upon the VGG-16 model~\cite{simonyan2014very} that contains 5 convolutional blocks. We drop the last pooling layer and the fully-connected layers to better fit for our task. Let us denote the outputs of each block, conv1\_2, conv2\_2, conv3\_3, conv4\_3, conv5\_3, respectively, by $\mathbf{A}_1$, $\mathbf{A}_2$, $\mathbf{A}_3$, $\mathbf{A}_4$, $\mathbf{A}_5$. Each block also produces a side output $\mathbf{F}_i$ by feeding $\mathbf{A}_i$ into two extra convolutional layers and an up-sampling layer. The feature aggregation strategy progressively fuses the feature $\mathbf{F}_i$ at scale $i$ with the fused feature $\tilde{\mathbf{F}}_{i+1}$ from scale $i+1$. In the end, a saliency map $\mathbf{S}$ is predicted based on the aggregated feature $\tilde{\mathbf{F}}_{1}$. Mathematically, we formulate the procedures of feature extraction and saliency prediction as follows:
\begin{align}
& \mathbf{F}_i = u\big(g\big(g(\mathbf{A}_i)\big)\big) \ \ \ \ \ \ \ 1\leq i \leq 5, \\  
& \tilde{\mathbf{F}}_i  =
\left\{
\begin{aligned}
& g([\tilde{\mathbf{F}}_{i+1}, \mathbf{F}_i]) & & 1\leq i < 5\\ 
& \mathbf{F}_i & & i = 5,  \\
\end{aligned}
\right. \\
& \mathbf{S} = h(\mathbf{W}_s \ast \tilde{\mathbf{F}}_1 + \mathbf{b}_s),
\end{align}
in which $g(\cdot)$ denotes the operations that consists of a 64-channel convolutional layer followed by a non-linear activation function. The kernel size of the convolution is  $3\times3$ and the stride is 1. $u(\cdot)$ is an upsampling operation using bilinear interpolation. $[ \cdot,\cdot]$ represents a channel-wise concatenation. $\mathbf{W}_s$ and $\mathbf{b}_s$ are, respectively, the parameters of the $1\times 1$ kernel and the bias. $\ast$ represents the convolution operator and $h(\cdot)$ is the Sigmoid function.

This stream structure is applied to predict RGB saliency and depth saliency separately. The RGB saliency prediction stream takes a 3-channel color image as input while the depth stream inputs a 1-channel depth map. Except the inputs, these two streams share the same structure but with different parameter values. In addition, it needs to be mentioned that we drop the superscription $^{rgb}$ or $^{d}$ in Eq.(1-3) for notational convenience.

\subsection{Saliency Fusion Module}
In contrast to previous RGB-D saliency detection works~\cite{chen2019multi, peng2014rgbd} that fuse multi-modal predictions by element-wise addition or multiplication, we design a saliency fusion module that learns a switch map for adaptive fusion of the RGB saliency prediction $\mathbf{S}^{rgb}$ and the depth saliency predictions $\mathbf{S}^{d}$. This module first concatenates the last layer features of two streams and then goes through a convolutional layer to learn a switch map $\mathbf{SW}$. In the end, a fused saliency map $\mathbf{S}^{fused}$ is obtained. All operations in this module are formulated by:
\begin{align}
& \tilde{\mathbf{F}}^{sw} = g([\tilde{\mathbf{F}}_1^{rgb}, \tilde{\mathbf{F}}_1^{d}]), \\
& \mathbf{SW} = h(\mathbf{W}_{sw} \ast \tilde{\mathbf{F}}^{sw} + \mathbf{b}_{sw}), \\
& \mathbf{S}^{fused} = \mathbf{SW} \odot \mathbf{S}^{rgb} + (\mathbf{1} - \mathbf{SW}) \odot \mathbf{S}^{d},
\end{align}
where $\tilde{\mathbf{F}}^{sw}$ represents the 64-channel feature fusing two modalities. $\tilde{\mathbf{F}}_1^{rgb}$ and $\tilde{\mathbf{F}}_1^{d}$ are, respectively, the features at the last layer of the color and depth streams. $\mathbf{W}_{sw}$ and $\mathbf{b}_{sw}$ are the parameters of the convolutional layer. $\odot$ denotes the element-wise multiplication.

The design of the switch map is motivated by the observation mentioned in Sec.~\ref{sec:introduction}. That is, good detection results are achieved in most scenarios if the algorithm can automatically choose the predictions from either RGB or depth modality. To this end, we construct a pseudo ground truth switch map $\mathbf{Y}^{sw}$ to guide the learning of $\mathbf{SW}$. It is defined by
\begin{align}
\mathbf{Y}^{sw} = \mathbf{S}^{rgb}\odot \mathbf{Y} + (1 - \mathbf{S}^{rgb})\odot(1 - \mathbf{Y}) \label{eq:SW_label}.
\end{align}
$\mathbf{Y}^{sw}$ gets 1 if the RGB saliency prediction $\mathbf{S}^{rgb}$ and the ground truth $\mathbf{Y}$ are both salient or nonsalient, and 0 otherwise. It means that if $\mathbf{S}^{rgb}$ correctly identifies salient objects, then we choose the prediction from the RGB stream as the final result; otherwise, the prediction from the depth stream is chosen. 

In implementation, the switch map is a 1-channel image whose pixel values are assigned in $[0,1]$. Therefore, instead of alternatively choosing the prediction from one or the other modality, the switch map plays a role to adaptively weigh the RGB and depth predictions, and therefore the fused saliency map is a weighted sum of the two predictions.

\subsection{Loss Function}
During training, a set of samples $\mathcal{C} = \{(\mathbf{X}_i , \mathbf{D}_i, \mathbf{Y}_i)\}_{i=1}^N$ are given, in which $N$ is the total number of samples. $\mathbf{X}_i = \{x_{i, j}\}_{j=1}^T$ and $\mathbf{D}_i = \{d_{i, j}\}_{j=1}^T$ are a pair of RGB and depth images with $T$ pixels. $\mathbf{Y}_i = \{y_{i, j}\}_{j=1}^T$ is the corresponding binary ground truth saliency map, with $1$ denoting salient pixels and $0$ for the background. Our network is trained to generate an edge-preserving saliency map by learning a switch map and fusing two unimodal saliency predictions. Therefore, the loss function is designed to contain three terms: a saliency loss $\mathcal{L}_{sal}$, a switch loss $\mathcal{L}_{sw}$, as well as an edge-preserving loss $\mathcal{L}_{edge}$. That is,
\begin{align}
& \mathcal{L} = \mathcal{L}_{sal} + \mathcal{L}_{sw} + \mathcal{L}_{edge} \label{eq:sal_loss}.
\end{align}

\textbf{Saliency Loss}. There are three saliency maps produced in our network: $\mathbf{S}^{rgb}$, $\mathbf{S}^d$, and $\mathbf{S}^{fused}$. We use the ground truth to supervise each of them. A standard cross-entropy loss is adopted to compute the difference between predicted results and the ground truth. Therefore, The saliency loss is defined by
\begin{align}
\mathcal{L}_{sal} =\mathcal{L}_{sal}^{rgb} + \mathcal{L}_{sal}^{d} + \mathcal{L}_{sal}^{fused},
\end{align}
where
\begin{align}
\begin{split}
\mathcal{L}_{sal}^{m} = & -\sum_{i=1}^N\sum_{j=1}^T \big( y_{i,j} \log \mathbf{S}^{m}_{i,j} \\
& + (1-y_{i,j})\log (1 - \mathbf{S}^{m}_{i,j})\big) \label{eq:CEloss}. \\
\end{split} 
\end{align}
Here, the superscript $m$ denotes a modality that may be $rgb$, $d$, or $fused$. $\mathbf{S}^{m}_{i,j}$ represents the probability predicted by the modality $m$ for pixel $j$ in the $i$-th image to be salient.

\textbf{Switch Loss}. The switch map is supervised by the pseudo ground truth $\mathbf{Y}^{sw}$ constructed in Eq.(\ref{eq:SW_label}). We use the cross-entropy loss to penalize the learning of the switch map as well. The loss is defined by:
\begin{align}
\begin{split}
\mathcal{L}_{sw} = & -\sum_{i=1}^N\sum_{j=1}^T \big( y_{i,j}^{sw} \log \mathbf{SW}_{i,j} \\
&  + (1-y_{i,j}^{sw})\log (1 - \mathbf{SW}_{i,j}) \big) \label{eq:SWloss} \\
\end{split} 
\end{align}
where $y_{i,j}^{sw}$ is the $j$-th pixel of the pseudo ground truth switch map for the $i$-th image. $\mathbf{SW}_{i,j}$ represents the probability for the pixel to choose the RGB prediction $\mathbf{S}^{rgb}_{i,j}$. 

\textbf{Edge-preserving Loss}. The edge-preserving property has been considered in previous RGB saliency detection works~\cite{zhang2017amulet,wang2018rgb} to obtain sharp salient object boundaries and improve detection performance. In contrast to these works that used superpixel boundaries as constraints~\cite{wang2018rgb} or adopted short connections in network for boundary refinement~\cite{zhang2017amulet}, we formulate the edge-preserving constraint as a loss term supervising the fused saliency map. It is defined by
\begin{align}
\begin{split}
\mathcal{L}_{edge} & = \frac{1}{N}\sum_{i = 1}^{N}||\partial_x(\mathbf{S}^{fused}_{i}) - \partial_x(\mathbf{Y}_{i})||_2^2 \\
& + ||\partial_y(\mathbf{S}^{fused}_{i}) - \partial_y(\mathbf{Y}_{i})||_2^2,
\end{split} 
\end{align}
where $\partial_x(\cdot)$ and $\partial_y(\cdot)$ are gradients in horizontal and vertical direction respectively. This loss preserves edges by minimizing the differences between the edges in the fused saliency maps and those in the ground truth maps.

\subsection{Implementation Details}
Our approach is implemented based upon TensorFlow~\cite{abadi2016tensorflow}. We adopt the VGG-16 model~\cite{simonyan2014very} as the backbone for a fair comparison with previous works. All parameters except those in VGG-16 are initialized via Xavier~\cite{glorot2010understanding}. Our entire network is trained in an end-to-end manner using the aforementioned loss function. The loss is optimized by the Adam optimizer~\cite{kingma2014adam} with a batch size of 8 and a learning rate of $10^{-4}$. All input images are resized to the resolution of $224 \times 224$ for training and test. We conduct our experiments on a PC with a single NVIDIA GTX 1080Ti GPU. The test time for each RGB-D image pair takes only 0.03s.

\section{Experimental Results}
\subsection{Datasets}
To validate the proposed approach, we conduct a series of experiments on three publicly available datasets: NJUD~\cite{ju2014depth}, NLPR~\cite{peng2014rgbd}, and STEREO~\cite{niu2012leveraging}. The NJUD dataset~\cite{ju2014depth} contains 2003 binocular image pairs collected from Internet, 3D movies and photographs. NLPR~\cite{peng2014rgbd} consists of 1000 images captured by Kinect, covering a variety of indoor and outdoor scenes under different illumination conditions. STEREO~\cite{niu2012leveraging} provides the Web links for downloading stereoscopic images and a total of 797 pairs are gathered.

For a fair comparison to state-of-the-arts, we utilize the same data split as in~\cite{han2017cnns}. The training set contains 1400 samples from the NJUD dataset and 650 samples from NLPR. 100 image pairs from NJUD and 50 image pairs from NLPR are sampled to form the validation set. The test set consists of the remaining data in these two datasets, together with the full STEREO dataset. In addition, we augment the training set by flipping all training samples horizontally. 
 
\subsection{Evaluation Metrics}
We adopt the precision-recall (PR) curves, the F-measure score, and the mean absolute error (MAE) for performance evaluation. These metrics are widely used in saliency detection tasks. The PR curves are plotted by binarizing a predicted saliency map using 255 thresholds equally distributed in [0, 1] and comparing the binarized map with the ground truth. The F-measure is a weighted harmonic mean of the precision and recall, defined by
\begin{align}
F_\beta =  \frac{(1+\beta^2) \cdot Precision \cdot Recall}{\beta^2 \cdot Precision + Recall}.
\end{align}
As done in previous works~\cite{han2017cnns, chen2019multi, chen2018progressively}, $\beta^2$ is set to be 0.3 for emphasizing the importance of precision. We compare two kinds of F-measure scores, which are the maximum F-measure and the mean F-measure, respectively. The maximum F-measure is the highest score computed by the PR pairs in PR curves. The mean F-measure is computed by using an adaptive threshold that is set to be the sum of mean and standard deviation of each saliency map. The MAE~\cite{zhang2017amulet} measures the saliency detection accuracy by
\begin{align}
MAE = \frac{1}{T} \sum_{j=1}^{T}|\mathbf{S}_{j} - \mathbf{Y}_{j}|.
\end{align}

\subsection{Ablation Study}
We first conduct experiments to validate the effectiveness of the components in our proposed model. To this end, different settings are considered:1) the full model, denoted by `AF'; 2) the model without edge-preserving loss, denoted by `AF-Edge'; 3) the model without switch map and edge-preserving loss, denoted by `AF-Edge-SW'; In this model, we concatenate the features from two streams and feed them into a $1\times1$ convolutional layer to predict the fused saliency map directly. 4) the model containing only the RGB saliency prediction stream, denoted by `$\mathbf{S}^{rgb}$'; and 5) the one containing only the depth stream, denoted by `$\mathbf{S}^{d}$'. Table~\ref{table:ablation} reports the mean F-measure scores for these models on three datasets. 

\begin{table}[htbp]
\renewcommand\arraystretch{1.2}
\centering
\caption{The results for component analysis.}
\begin{tabular}{|p{50pt}|c|c|c|}
\hline
Settings & NJUD & NLPR & STEREO \\ \hline
$\mathbf{S}^{rgb}$ & 0.854 & 0.857 & 0.874 \\
$\mathbf{S}^d$ & 0.800 & 0.754 & 0.770 \\
AF-Edge-SW & 0.872 & 0.865 & 0.879 \\
AF-Edge & 0.874 & 0.873 & 0.886 \\
AF & \bf{0.878} & \bf{0.881} & \bf{0.891}\\ \hline
\end{tabular}
\label{table:ablation}
\end{table}

\subsubsection*{The effectiveness of the saliency fusion module}
The comparison of `AF-Edge-SW' and `AF-Edge' in Table~\ref{table:ablation} demonstrates the improvement in mean F-measure with our saliency fusion module. The results in Fig.~\ref{fig:SW} illustrate the fusion of $\mathbf{S}^{rgb}$ and $\mathbf{S}^{d}$ visually. When $\mathbf{S}^{rgb}$ correctly detects the salient objects, as the scenarios shown in the first two rows, our approach fuses more information from the RGB predictions by highlighting most regions in the switch maps. When objects share similar color appearances with backgrounds but have different depth values, as shown in the third row, our approach suppresses unreliable predictions in $\mathbf{S}^{rgb}$ by assigning low weights for these regions in the switch map. Thus, more information from $\mathbf{S}^{d}$ are fused. As expected, the proposed saliency fusion module can tackle these three types of scenarios correctly. 

\begin{figure}[htbp] 
\centering
\subfigure[RGB]{
\centering
\begin{minipage}{0.19\linewidth}
\centering
\includegraphics[width=1\columnwidth]{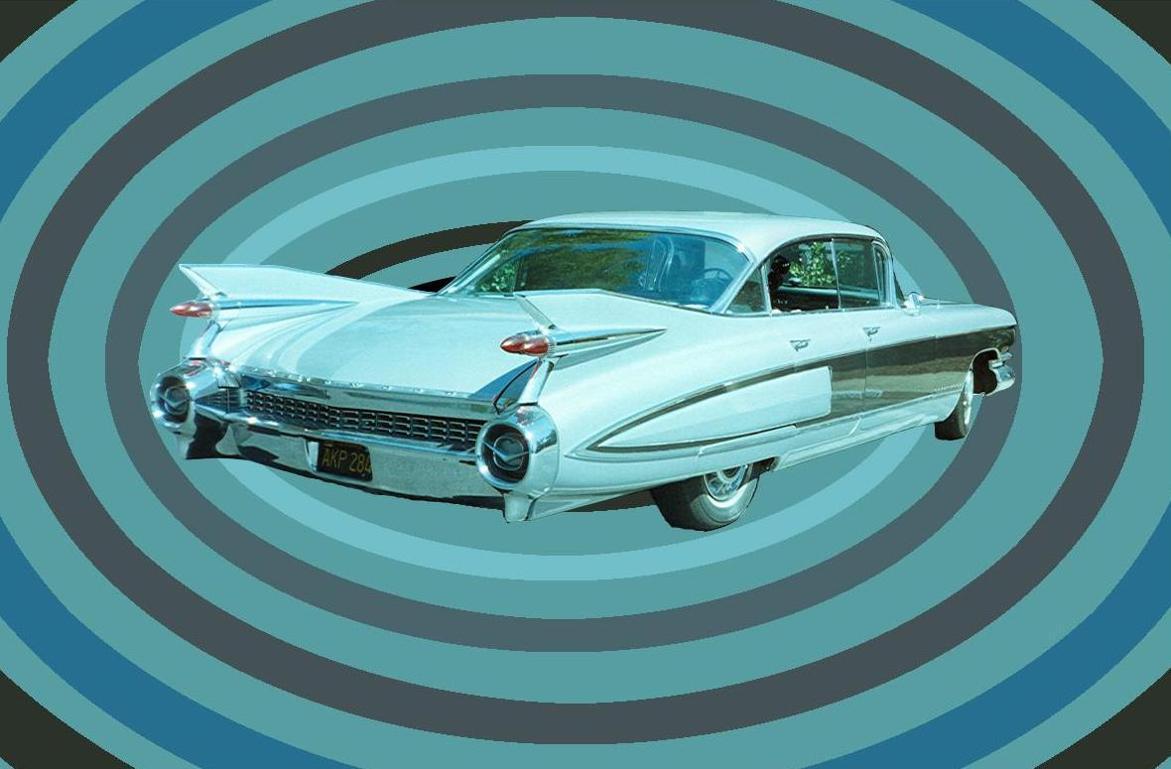} \\\vspace{1.5pt}
\includegraphics[width=1\columnwidth]{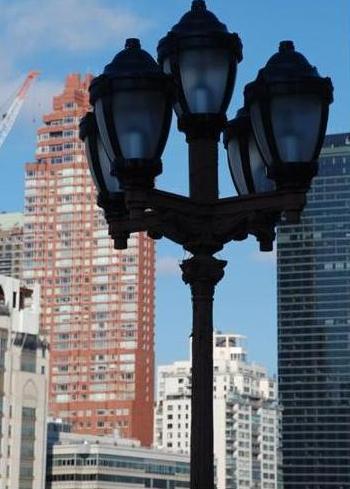} \\\vspace{1.5pt}
\includegraphics[width=1\columnwidth]{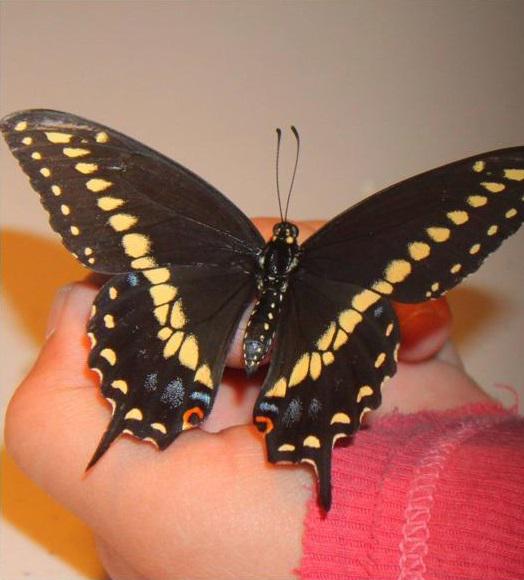} \\\vspace{1.5pt}
\includegraphics[width=1\columnwidth]{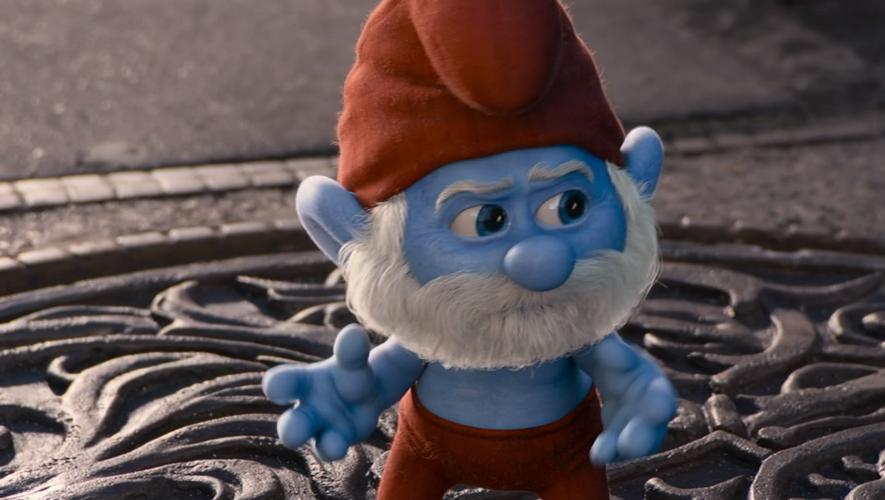} \\\vspace{1.5pt}
\includegraphics[width=1\columnwidth]{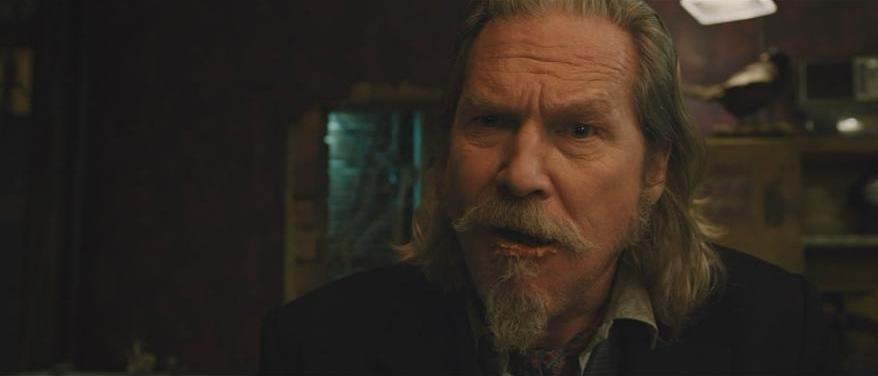} \\\vspace{1.5pt}
\end{minipage} 
}
\hspace{-12pt}
\subfigure[Depth]{
\centering
\begin{minipage}{0.19\linewidth}
\centering
\includegraphics[width=1\columnwidth]{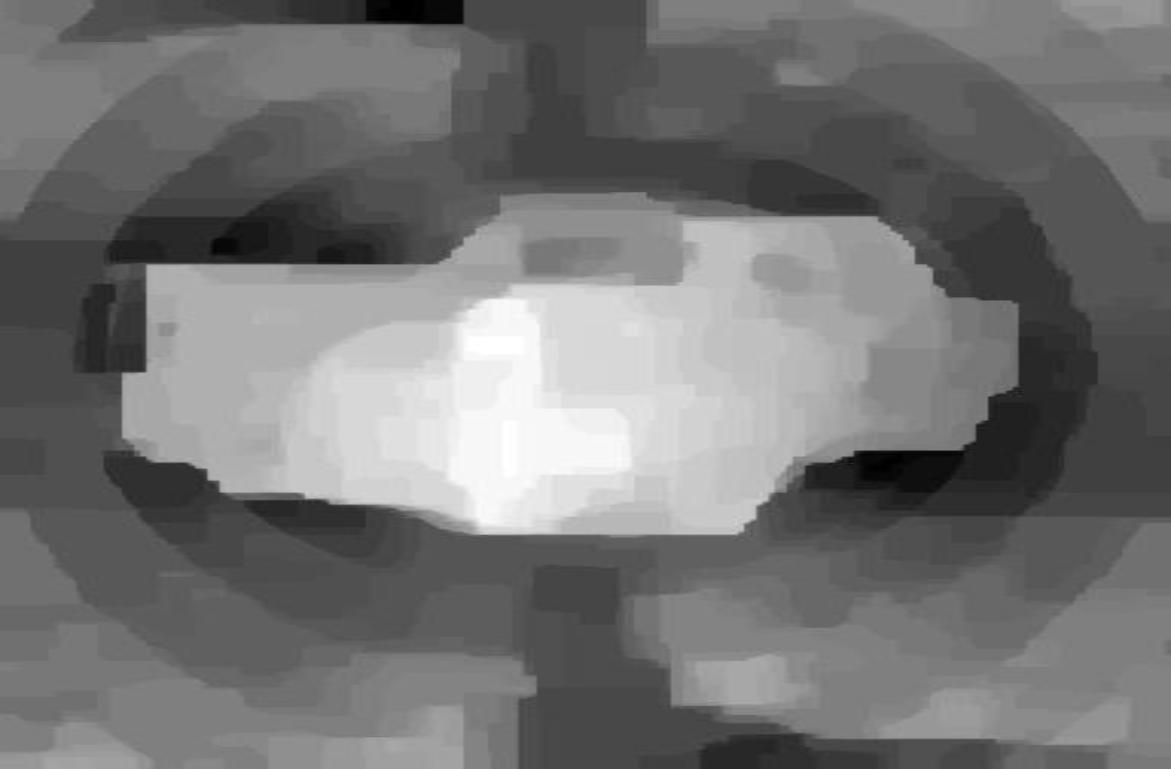} \\\vspace{1.5pt}
\includegraphics[width=1\columnwidth]{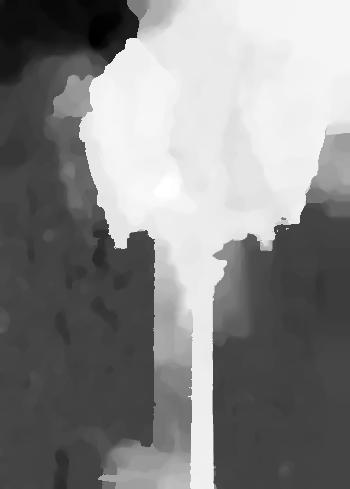} \\\vspace{1.5pt}
\includegraphics[width=1\columnwidth]{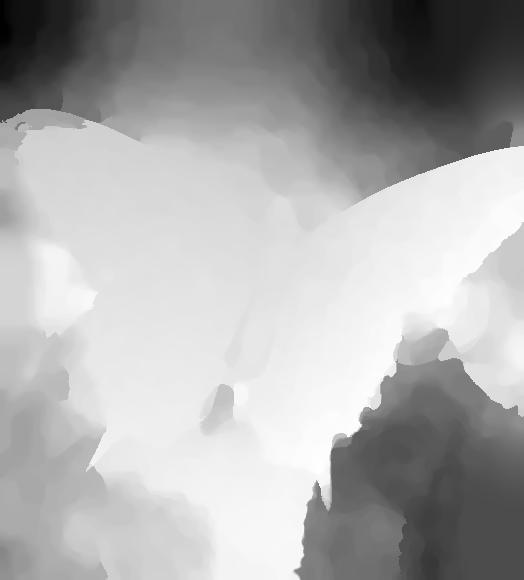} \\\vspace{1.5pt}
\includegraphics[width=1\columnwidth]{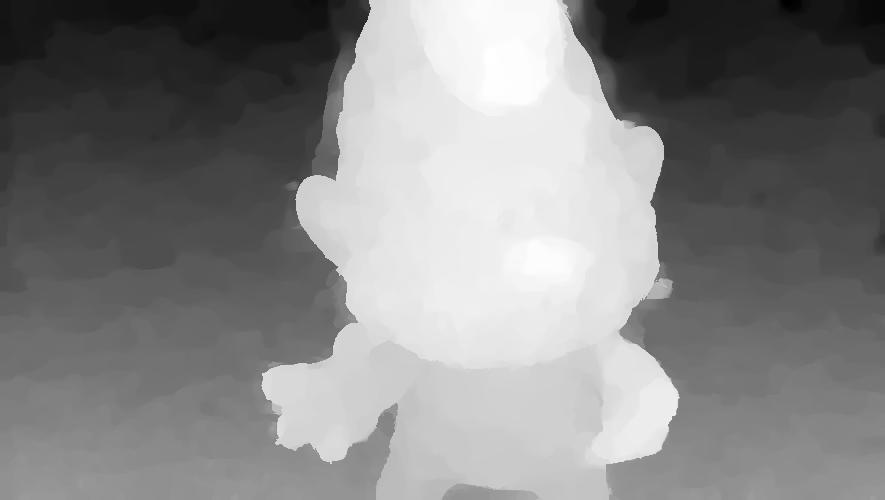} \\\vspace{1.5pt}
\includegraphics[width=1\columnwidth]{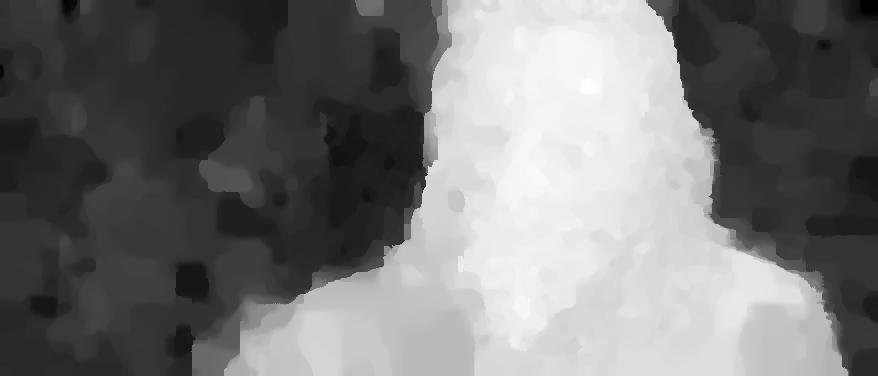} \\\vspace{1.5pt}
\end{minipage} 
}
\hspace{-12pt}
\subfigure[AF-Edge]{
\centering
\begin{minipage}{0.19\linewidth}
\centering
\includegraphics[width=1\columnwidth]{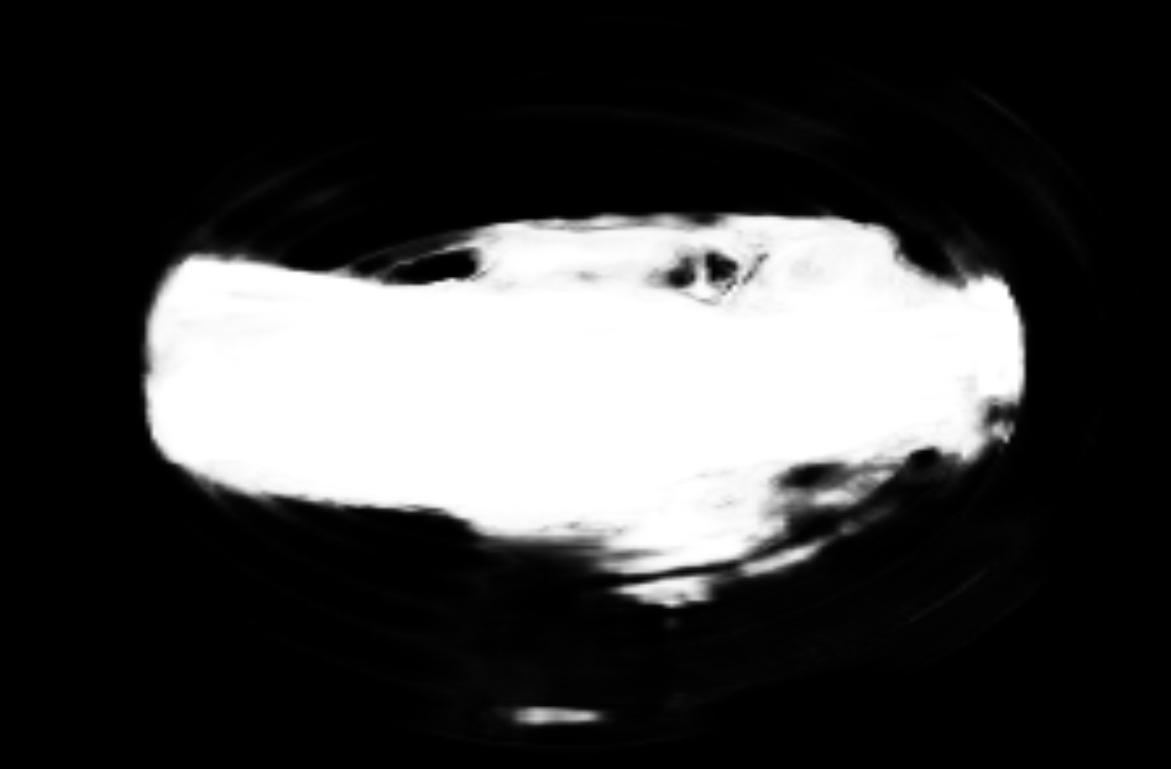} \\\vspace{1.5pt}
\includegraphics[width=1\columnwidth]{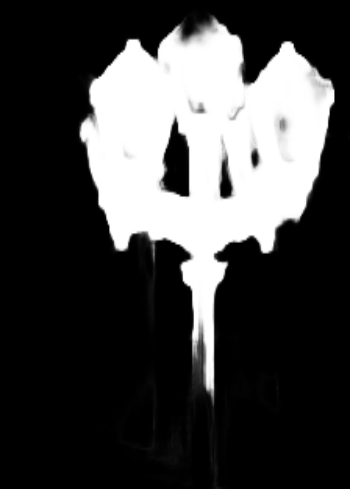} \\\vspace{1.5pt}
\includegraphics[width=1\columnwidth]{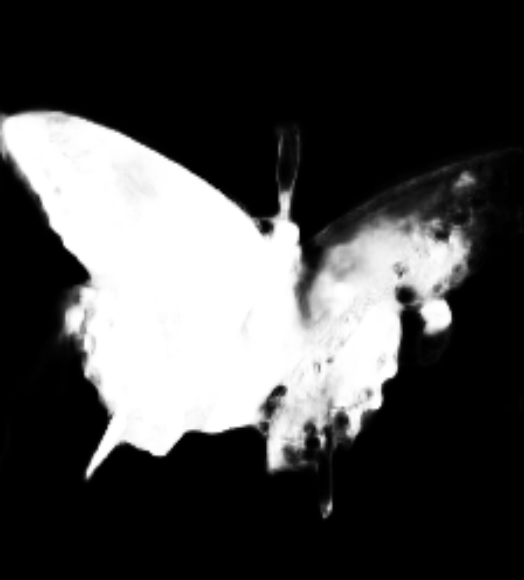} \\\vspace{1.5pt}
\includegraphics[width=1\columnwidth]{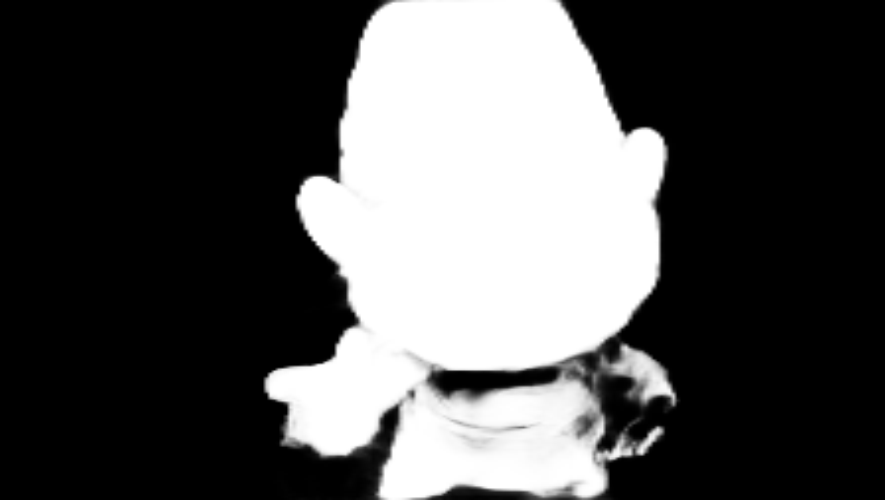} \\\vspace{1.5pt}
\includegraphics[width=1\columnwidth]{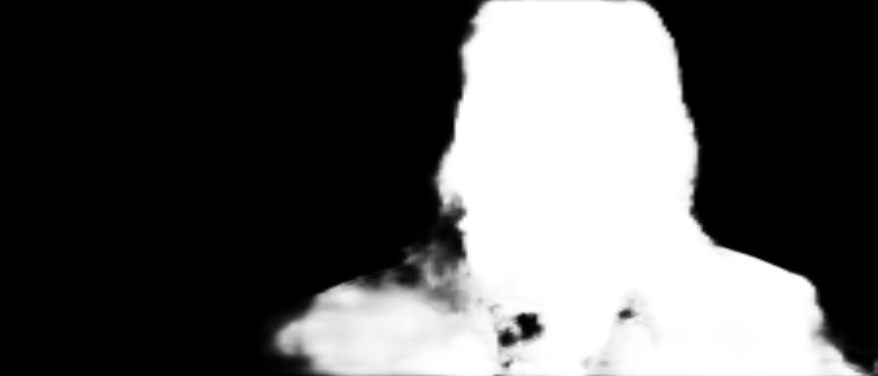} \\\vspace{1.5pt}
\end{minipage} 
}
\hspace{-12pt}
\subfigure[AF]{
\centering
\begin{minipage}{0.19\linewidth}
\centering
\includegraphics[width=1\columnwidth]{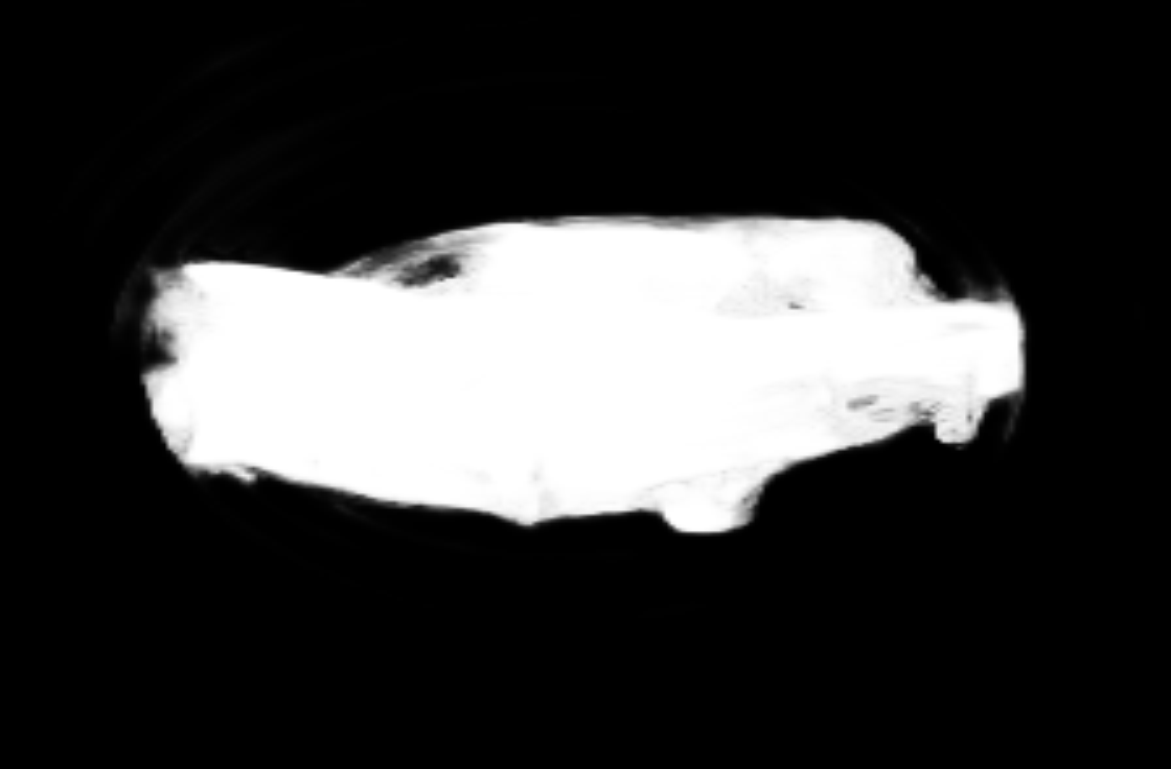} \\\vspace{1.5pt}
\includegraphics[width=1\columnwidth]{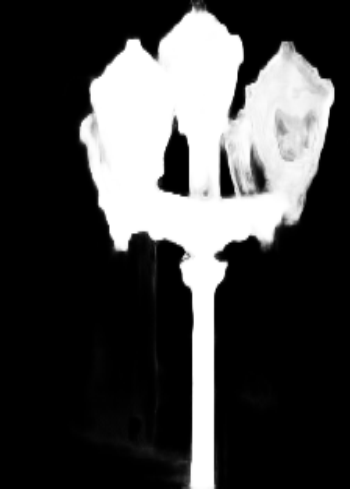} \\\vspace{1.5pt}
\includegraphics[width=1\columnwidth]{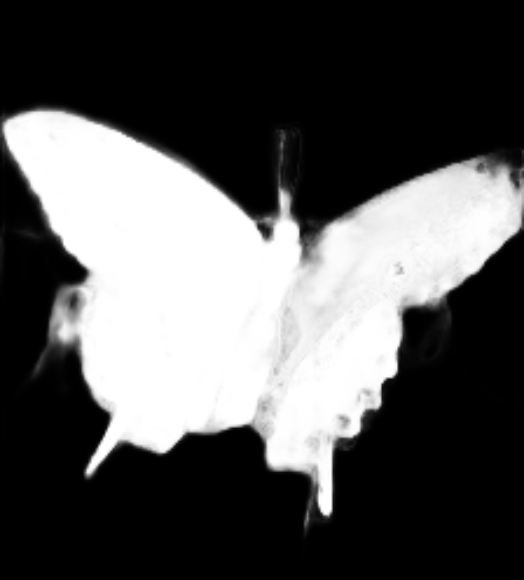} \\\vspace{1.5pt}
\includegraphics[width=1\columnwidth]{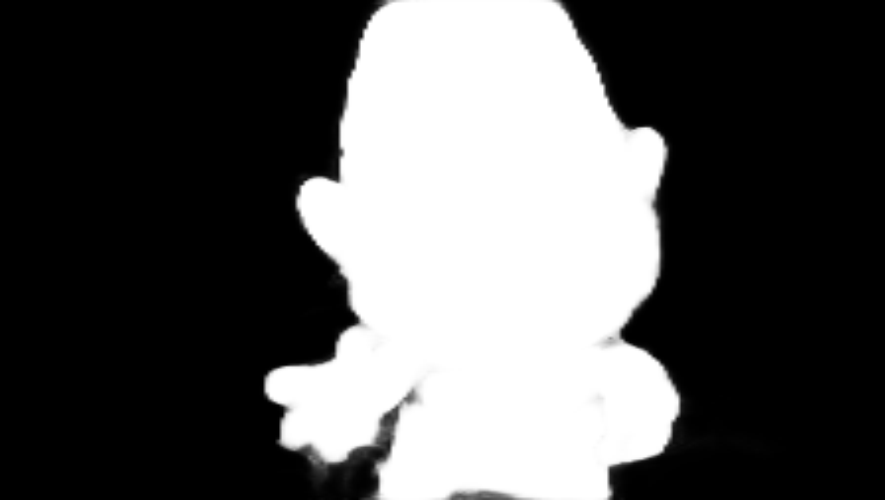} \\\vspace{1.5pt}
\includegraphics[width=1\columnwidth]{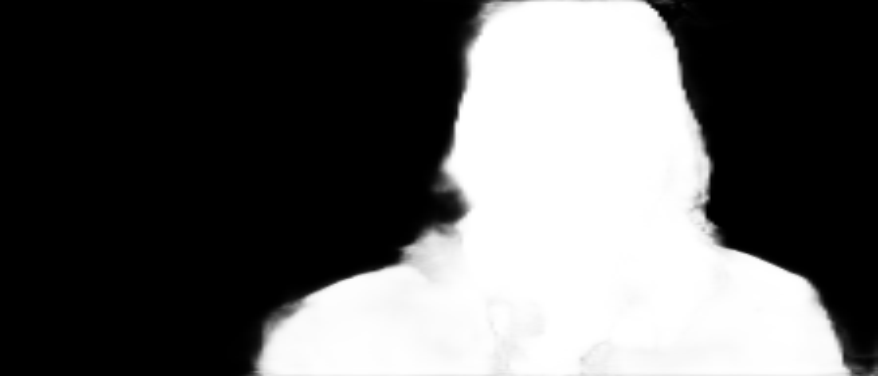} \\\vspace{1.5pt}
\end{minipage} 
}
\hspace{-12pt}
\subfigure[GT]{
\centering
\begin{minipage}{0.19\linewidth}
\centering
\includegraphics[width=1\columnwidth]{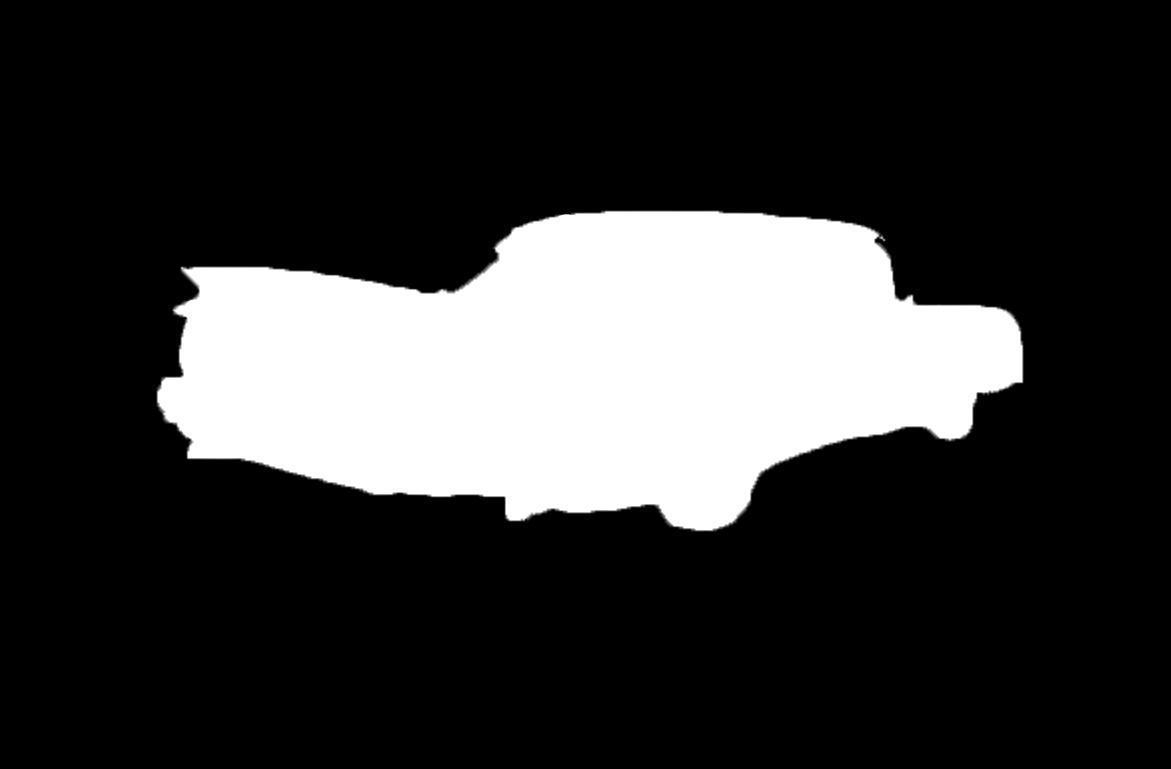} \\\vspace{1.5pt}
\includegraphics[width=1\columnwidth]{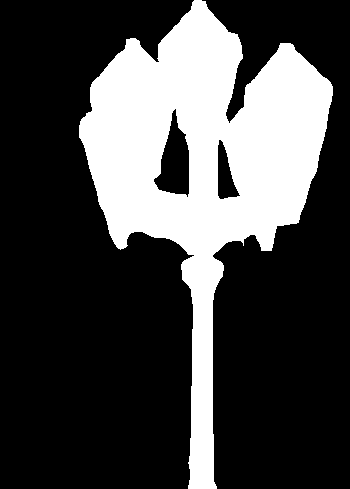} \\\vspace{1.5pt}
\includegraphics[width=1\columnwidth]{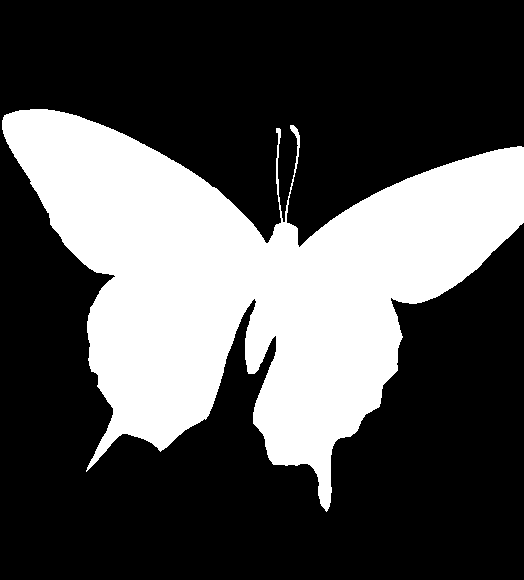} \\\vspace{1.5pt}
\includegraphics[width=1\columnwidth]{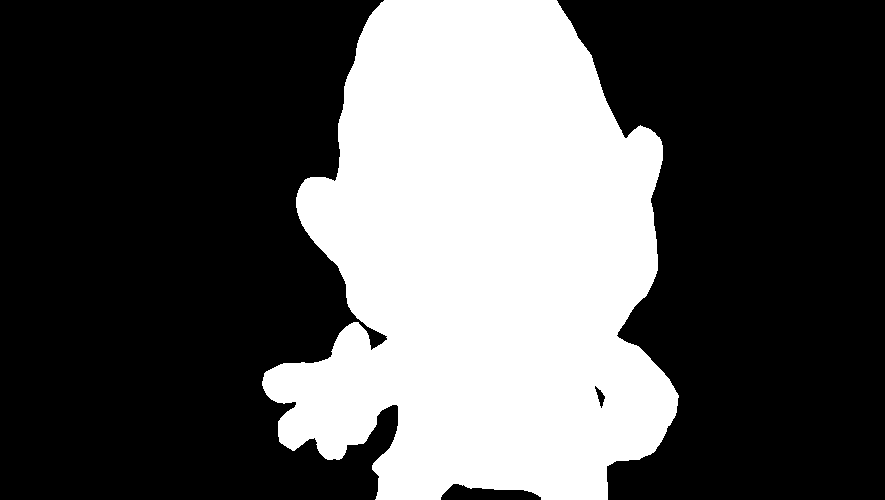} \\\vspace{1.5pt}
\includegraphics[width=1\columnwidth]{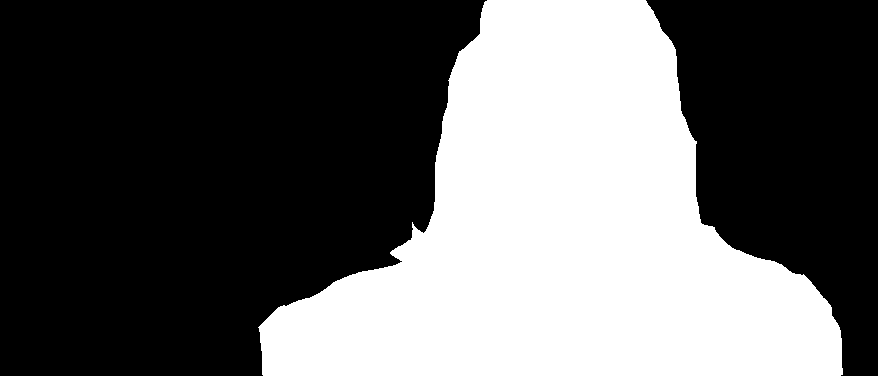} \\\vspace{1.5pt}
\end{minipage} 
}
\caption{Comparison of predictions with and without the edge-preserving loss.}     
\label{fig:edge}     
\end{figure}

\subsubsection*{The effectiveness of the edge-preserving loss}
With the edge-preserving loss, `AF' achieves superior performance to `AF-Edge' as reported in Table~\ref{table:ablation}. The results in Fig.~\ref{fig:edge} illustrate that the saliency maps predicted by `AF' can reduce the blur effect around objects' boundaries when the objects have similar appearances with the background. In addition, the salient objects are detected more coherently and completely with the edge-preserving constraint. The superiority in both quantitative and qualitative comparisons proves the effectiveness of this loss.

\subsection{Comparison with the State-of-the-arts}
We further compare our full model with two traditional methods including GP~\cite{ren2015exploiting} and LBE~\cite{feng2016local}, together with three CNN-based RGB-D saliency detection networks, including CTMF~\cite{han2017cnns}, MPCI~\cite{chen2019multi} and PCA~\cite{chen2018progressively}. The quantitative comparisons are reported in Table~\ref{table:maxF_MAE}, Fig.~\ref{fig:F-measure}, and Fig.~\ref{fig:PRcurves}. Qualitative comparisons are demonstrated in Fig.~\ref{fig:results}.

\subsubsection*{Quantitative Comparison}
\begin{figure}[htbp] 
\centering
\includegraphics[width=1\columnwidth]{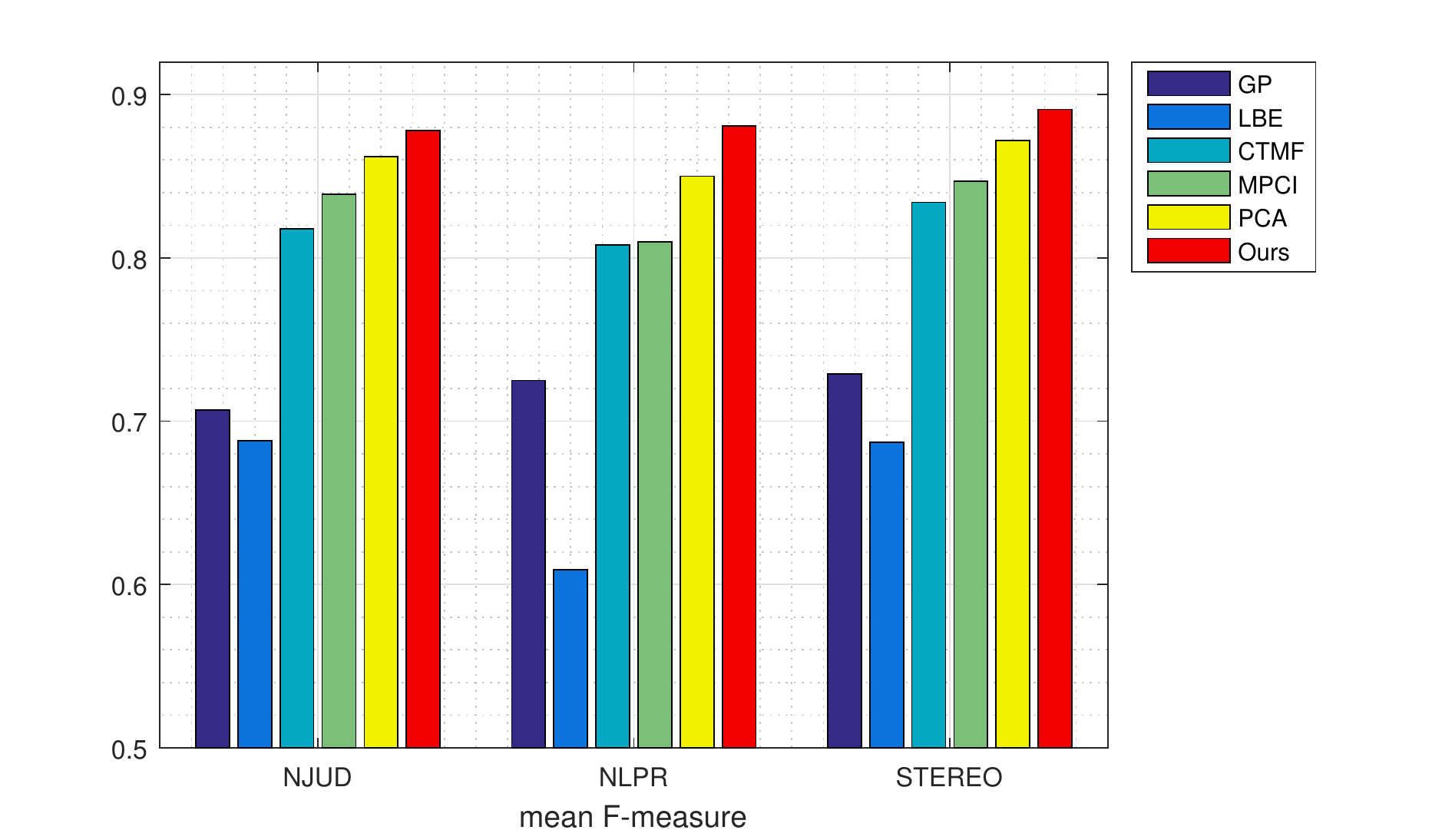}    
\caption{Comparison of mean F-measure.}     
\label{fig:F-measure}     
\end{figure}

\begin{figure*}[htbp] 
\centering
\subfigure[NJUD]{
\includegraphics[width=0.66\columnwidth]{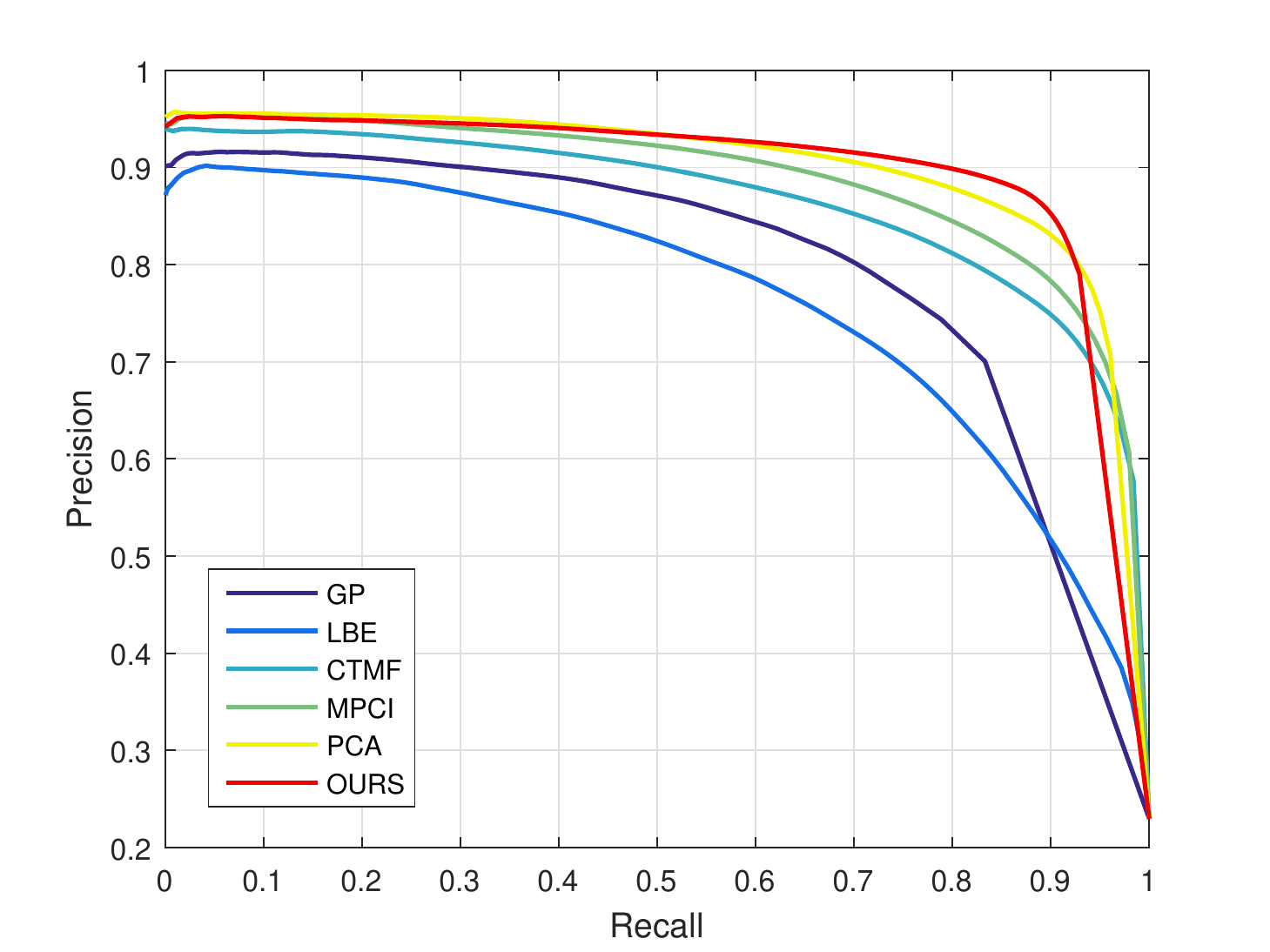}    
}
\hspace{-0.8cm}
\subfigure[NLPR]{
\includegraphics[width=0.66\columnwidth]{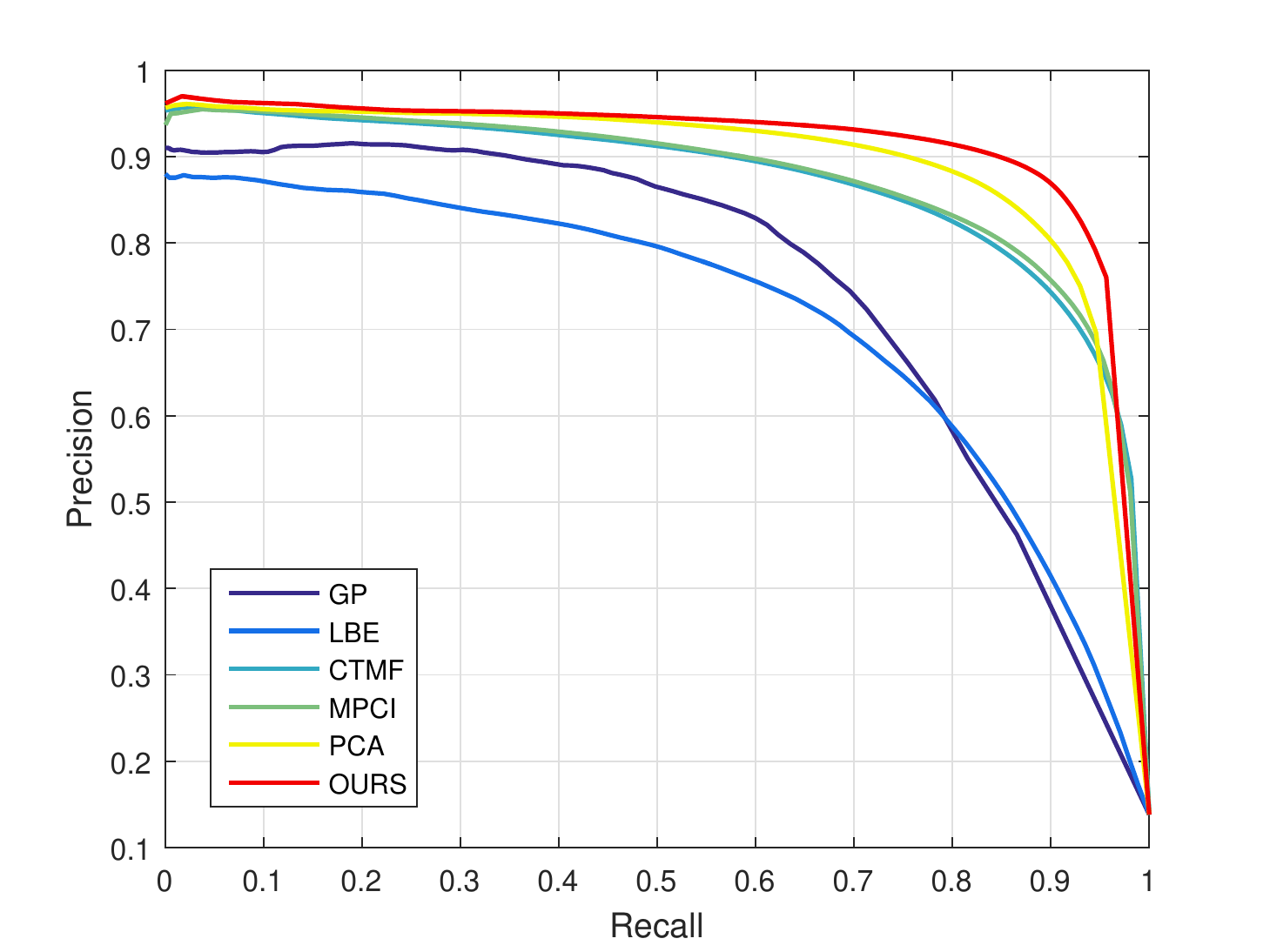}    
}
\hspace{-0.8cm}
\subfigure[STEREO]{
\includegraphics[width=0.66\columnwidth]{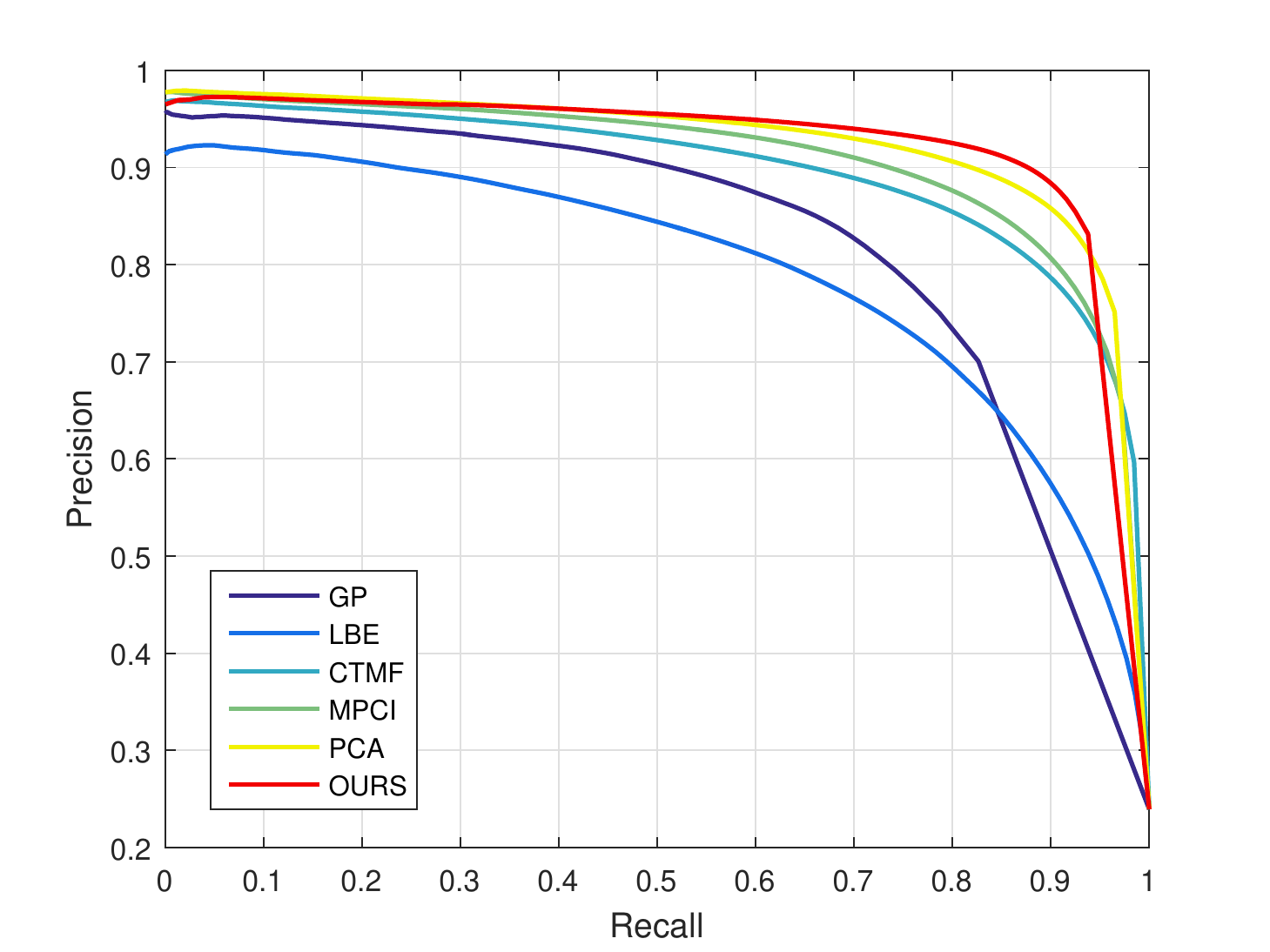}    
}
\caption{Comparison of PR curves.}     
\label{fig:PRcurves}     
\end{figure*}

As shown in Table~\ref{table:maxF_MAE}, Fig.~\ref{fig:F-measure}, and Fig.~\ref{fig:PRcurves}, the proposed method outperforms other state-of-the-art methods in terms of all evaluation metrics. Table~\ref{table:maxF_MAE} and Fig.~\ref {fig:F-measure} show that all deep learning based approaches outperform traditional methods by a great margin; and end-to-end frameworks, including PCA~\cite{chen2018progressively} and our approach, are superior to multi-stage methods such as CTMF~\cite{han2017cnns} and MPCI~\cite{chen2019multi}. Moreover, benefited from our fusion scheme and edge-preserving loss, the proposed method consistently improves the F-measure and MAE achieved by PCA on all three datasets, especially on NLPR where accurate depth data are collected by Kinect. The results indicate that our model can fuse depth information with RGB data more effectively.

\begin{table}[htb]
\centering
\caption{Comparison of maximum F-measure and MAE.}
\label{table:maxF_MAE}
\renewcommand\arraystretch{1.2}
\begin{tabular}{{|p{30pt}|c|c|c|c|c|c|c|}}  \hline 
               & \multicolumn{2}{c|}{NJUD} & \multicolumn{2}{c|}{NLPR} & \multicolumn{2}{c|}{STEREO}  \\ \hline
Methods & $F_\beta$ & MAE & $F_\beta$ & MAE  & $F_\beta$ & MAE  \\ \hline
GP          & 0.773 & 0.1679 & 0.764 & 0.1108 & 0.783 & 0.1564  \\ 
LBE        & 0.718 & 0.2381 & 0.687 & 0.2191 & 0.698 & 0.2421  \\ 
CTMF     & 0.857 & 0.0847 & 0.841 & 0.0554 & 0.853 & 0.0849  \\
MPCI      & 0.868 & 0.0789 & 0.841 & 0.0585 & 0.861 & 0.0780  \\ 
PCA        & 0.887 & 0.0592 & 0.864 & 0.0433 & 0.884 & 0.0592  \\ 
AF          & \textbf{0.899} & \textbf{0.0534} & \textbf{0.899} & \textbf{0.0327} & \textbf{0.904} & \textbf{0.0462}   \\ \hline
\end{tabular}
\end{table}

\subsubsection*{Qualitative Comparison}
\begin{figure*}[htbp] 
\centering
\subfigure[RGB]{
\centering
\begin{minipage}{0.1\linewidth}
\centering
\includegraphics[width=1\columnwidth]{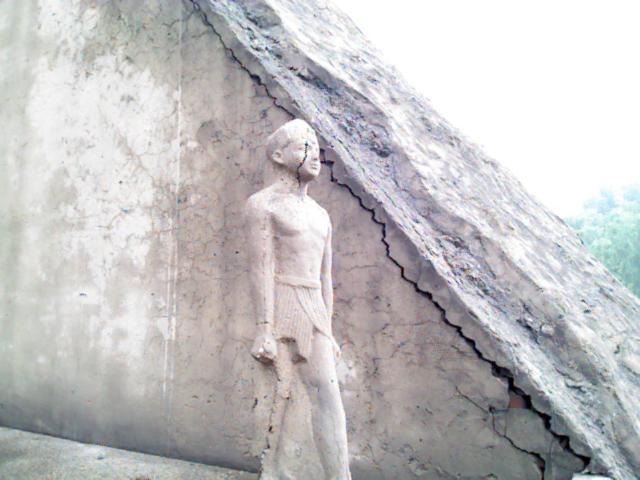} \\\vspace{1.5pt}
\includegraphics[width=1\columnwidth]{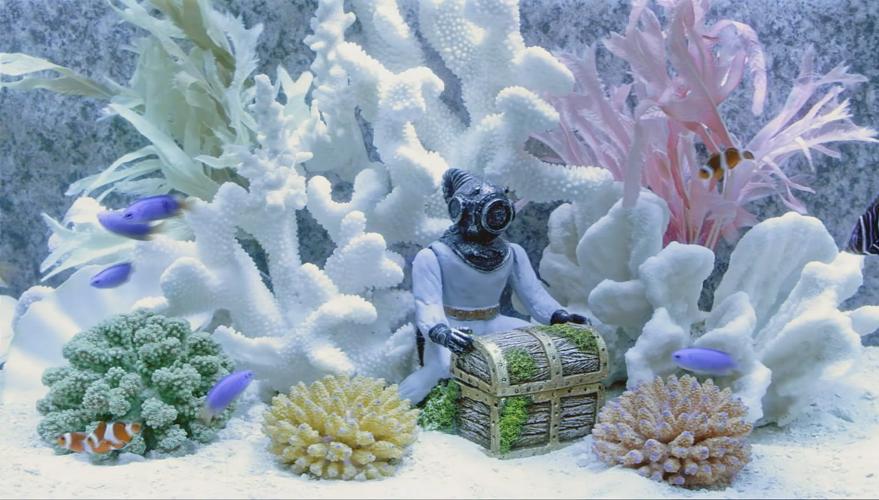} \\\vspace{1.5pt}
\includegraphics[width=1\columnwidth]{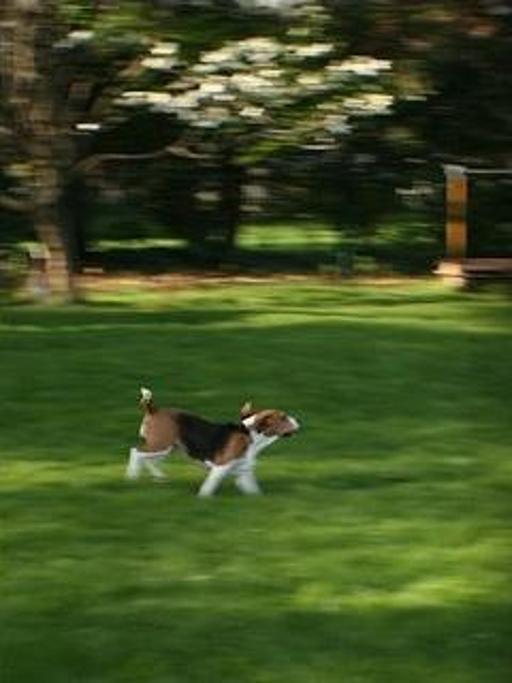} \\\vspace{1.5pt}
\includegraphics[width=1\columnwidth]{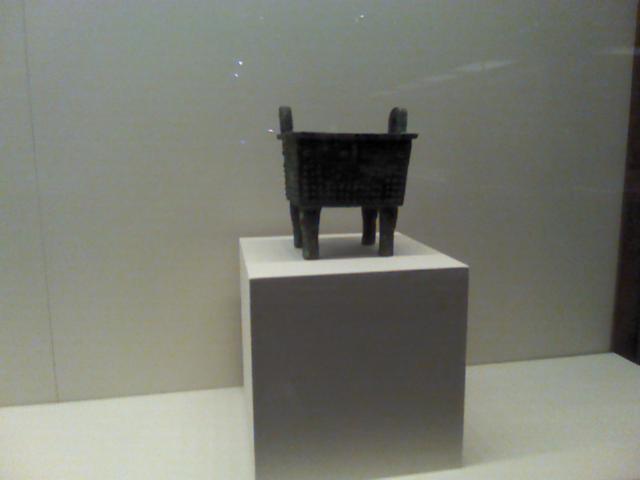} \\\vspace{1.5pt}
\includegraphics[width=1\columnwidth]{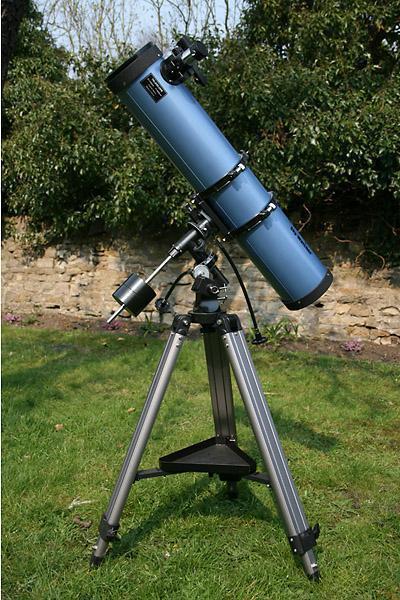} \\\vspace{1.5pt}
\includegraphics[width=1\columnwidth]{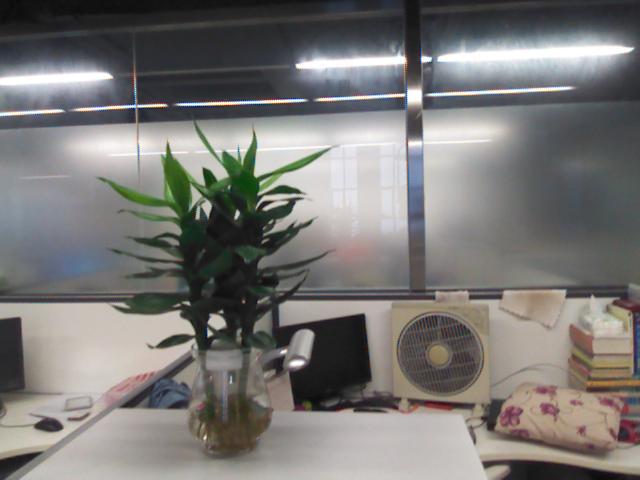} \\\vspace{1.5pt}
\end{minipage} 
}
\hspace{-12pt}
\subfigure[Depth]{
\centering
\begin{minipage}{0.1\linewidth}
\centering
\includegraphics[width=1\columnwidth]{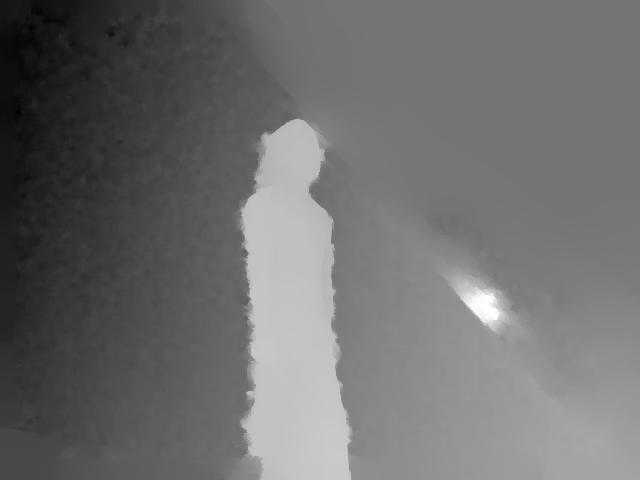} \\\vspace{1.5pt}
\includegraphics[width=1\columnwidth]{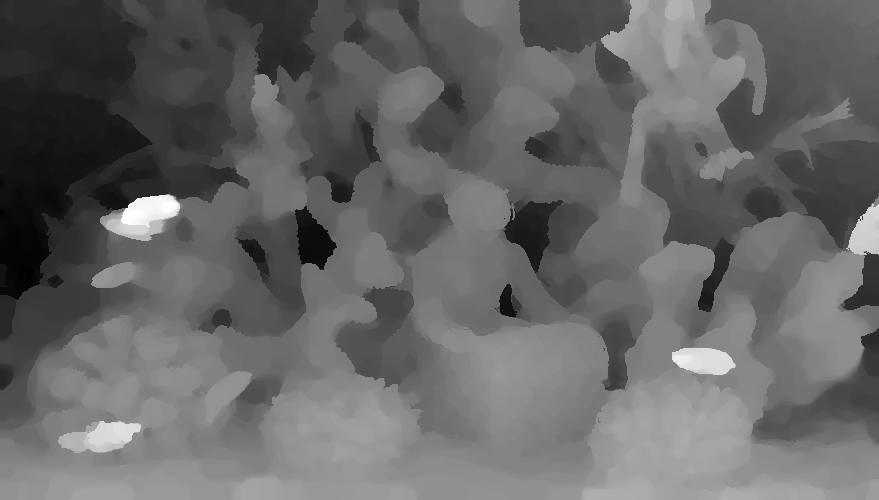} \\\vspace{1.5pt}
\includegraphics[width=1\columnwidth]{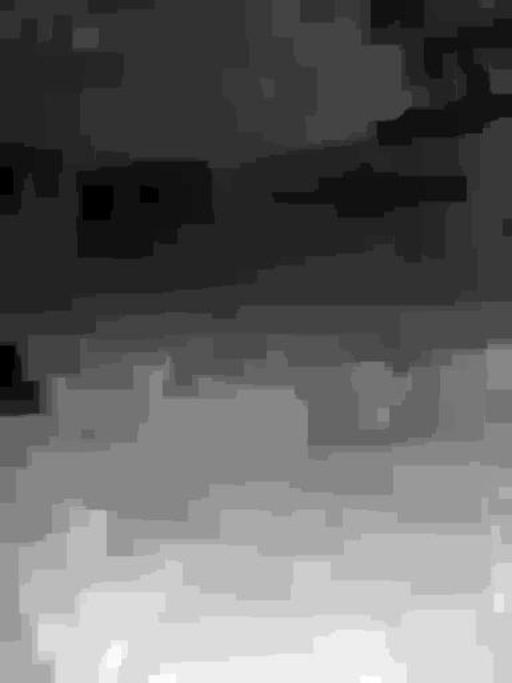} \\\vspace{1.5pt}
\includegraphics[width=1\columnwidth]{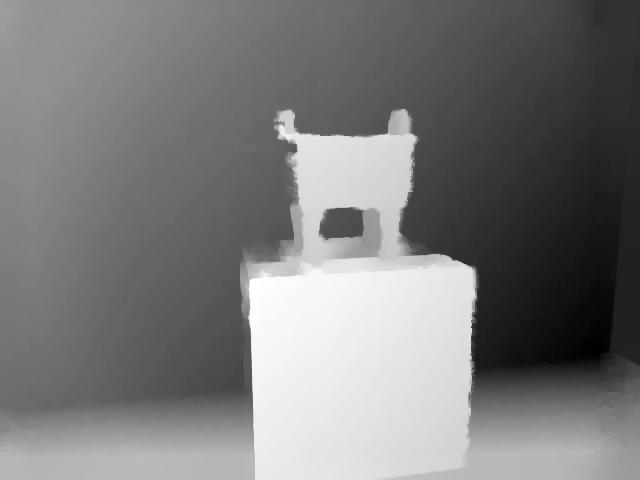} \\\vspace{1.5pt}
\includegraphics[width=1\columnwidth]{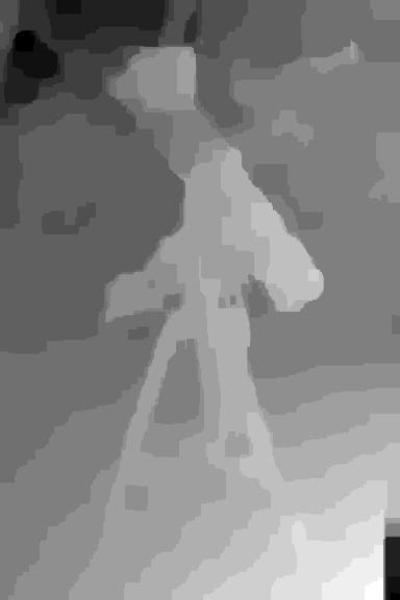} \\\vspace{1.5pt}
\includegraphics[width=1\columnwidth]{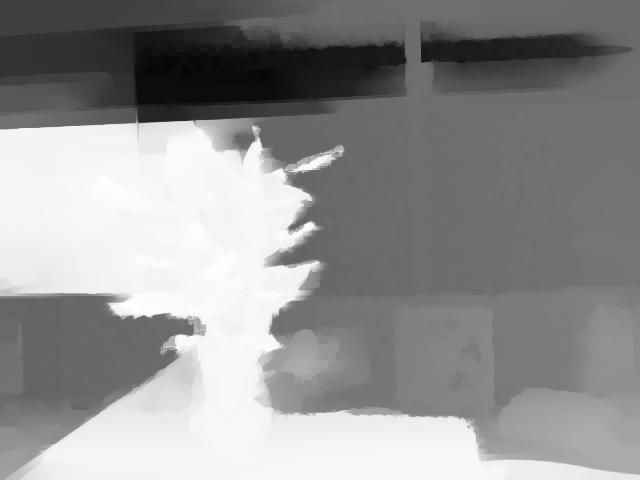} \\\vspace{1.5pt}
\end{minipage} 
}
\hspace{-12pt}
\subfigure[GP]{
\centering
\begin{minipage}{0.1\linewidth}
\centering
\includegraphics[width=1\columnwidth]{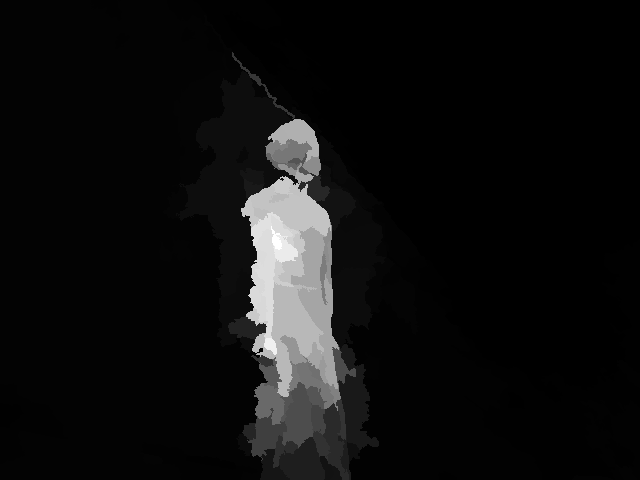} \\\vspace{1.5pt}
\includegraphics[width=1\columnwidth]{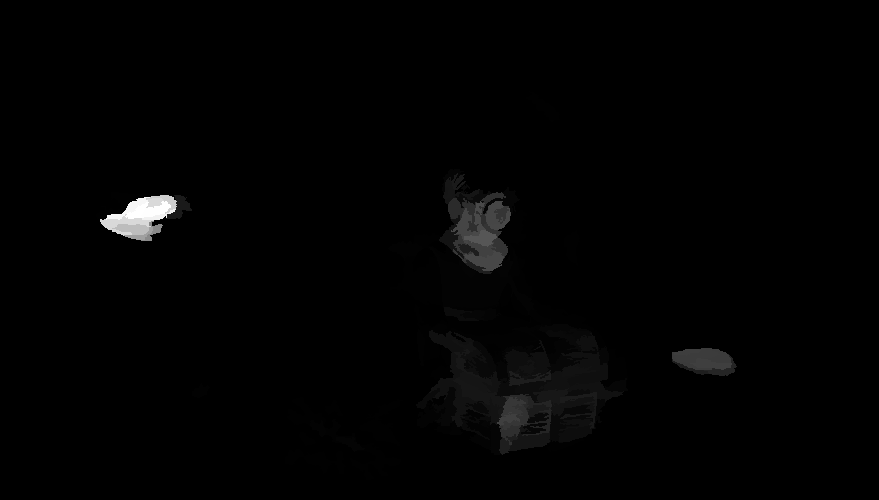} \\\vspace{1.5pt}
\includegraphics[width=1\columnwidth]{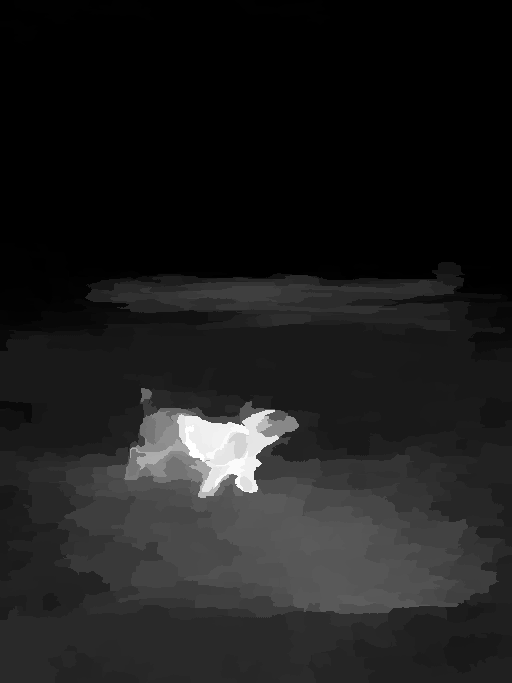} \\\vspace{1.5pt}
\includegraphics[width=1\columnwidth]{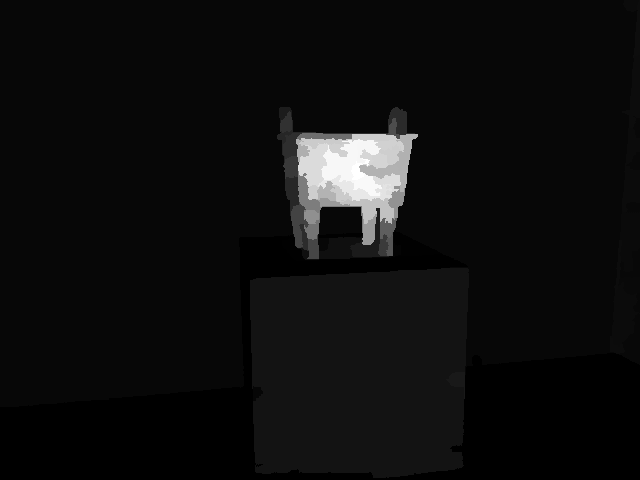} \\\vspace{1.5pt}
\includegraphics[width=1\columnwidth]{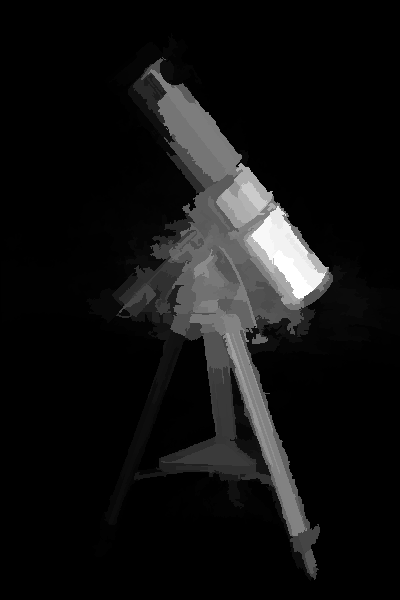} \\\vspace{1.5pt}
\includegraphics[width=1\columnwidth]{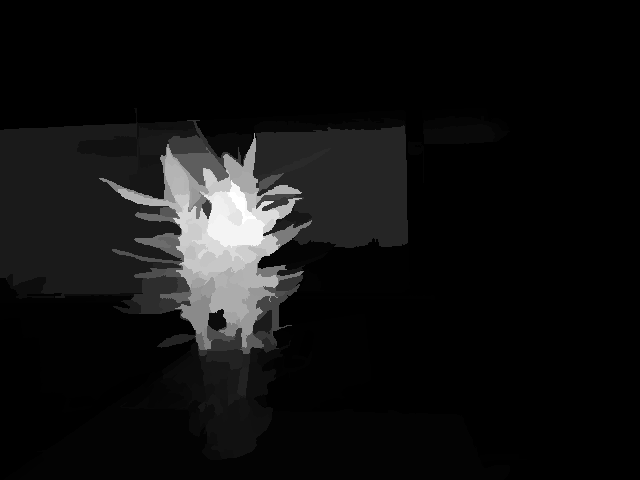} \\\vspace{1.5pt}
\end{minipage}
}
\hspace{-12pt}
\subfigure[LBE]{
\centering
\begin{minipage}{0.1\linewidth}
\centering
\includegraphics[width=1\columnwidth]{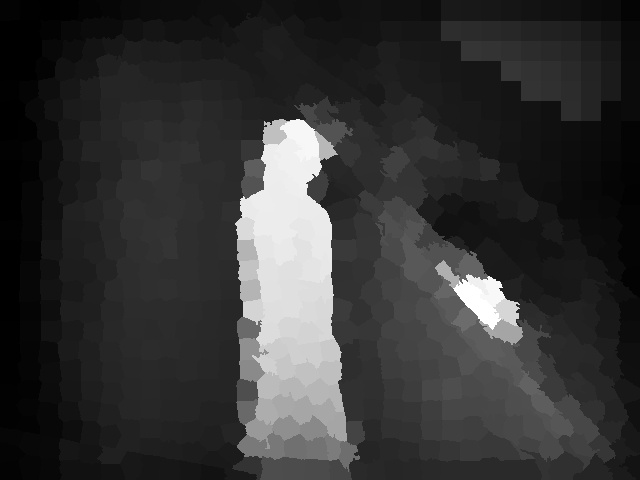} \\\vspace{1.5pt}
\includegraphics[width=1\columnwidth]{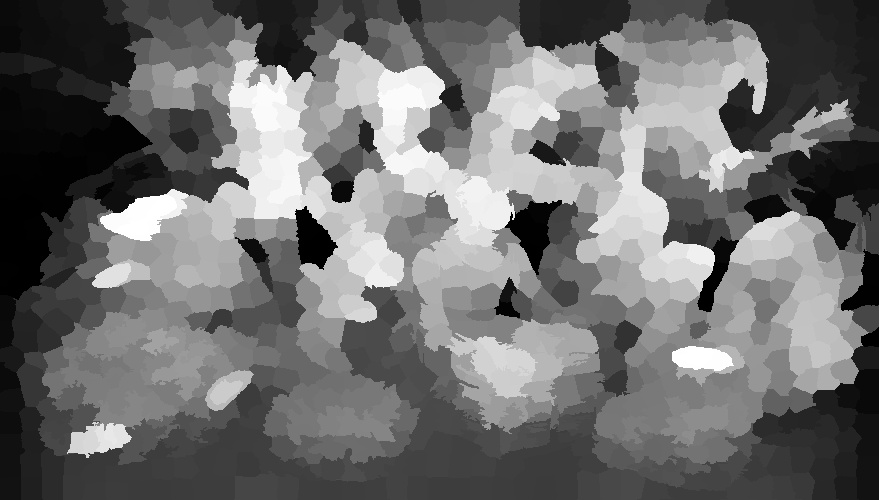} \\ \vspace{1.5pt}
\includegraphics[width=1\columnwidth]{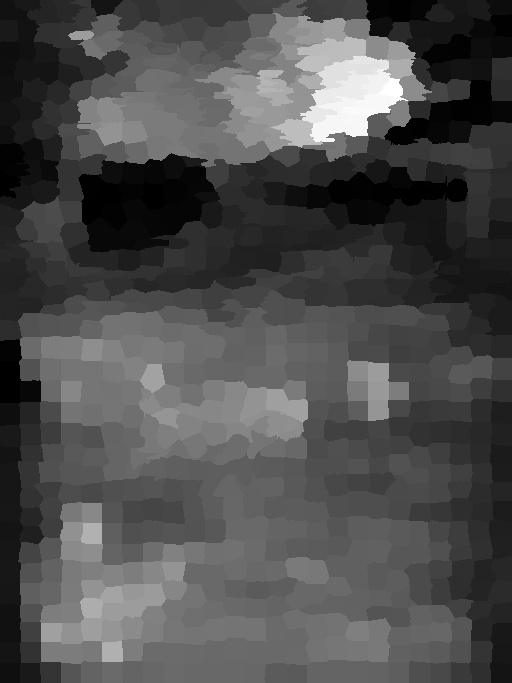} \\\vspace{1.5pt}
\includegraphics[width=1\columnwidth]{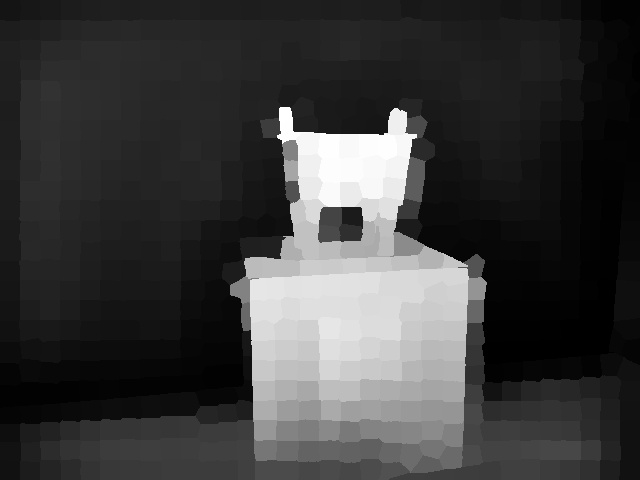} \\ \vspace{1.5pt}
\includegraphics[width=1\columnwidth]{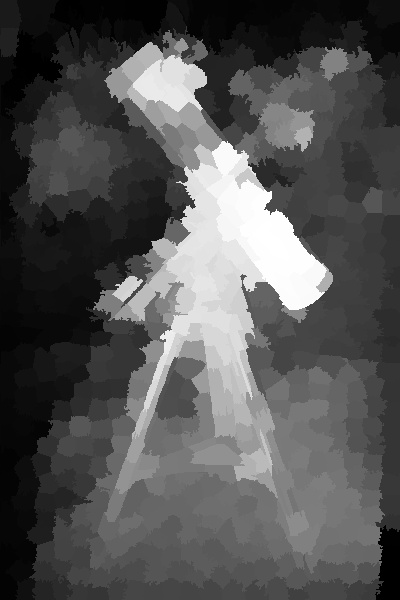} \\\vspace{1.5pt}
\includegraphics[width=1\columnwidth]{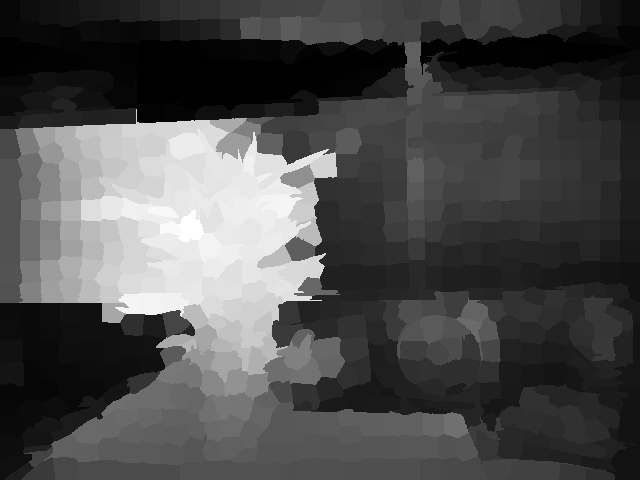} \\ \vspace{1.5pt}
\end{minipage}  
}
\hspace{-12pt}
\subfigure[CTMF]{
\centering
\begin{minipage}{0.1\linewidth}
\centering
\includegraphics[width=1\columnwidth]{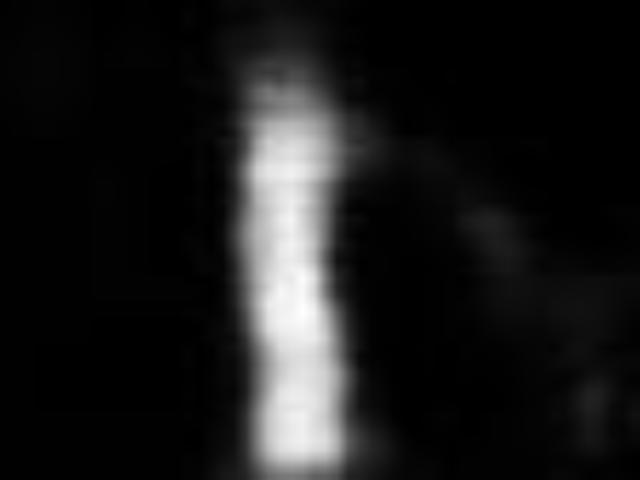} \\\vspace{1.5pt}
\includegraphics[width=1\columnwidth]{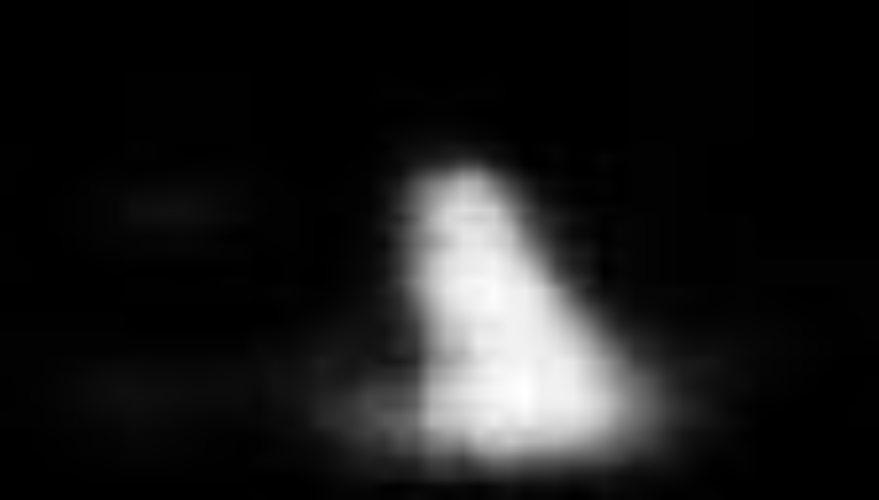} \\\vspace{1.5pt}
\includegraphics[width=1\columnwidth]{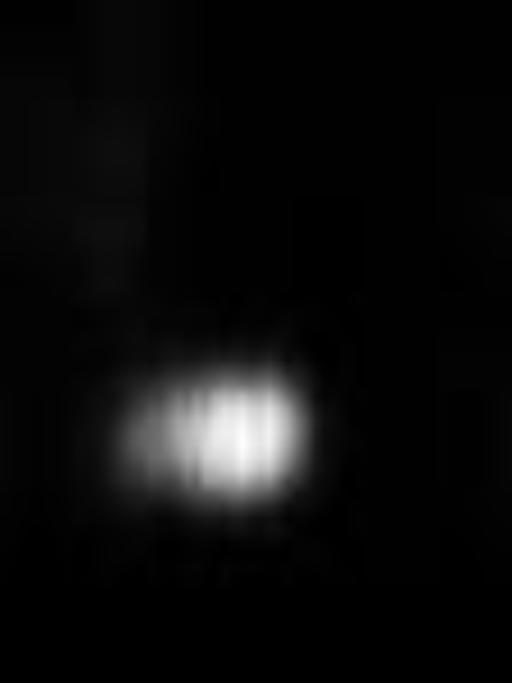} \\\vspace{1.5pt}
\includegraphics[width=1\columnwidth]{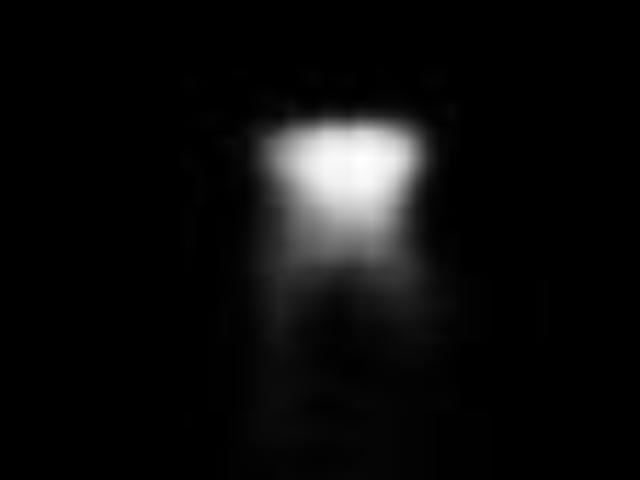} \\\vspace{1.5pt}
\includegraphics[width=1\columnwidth]{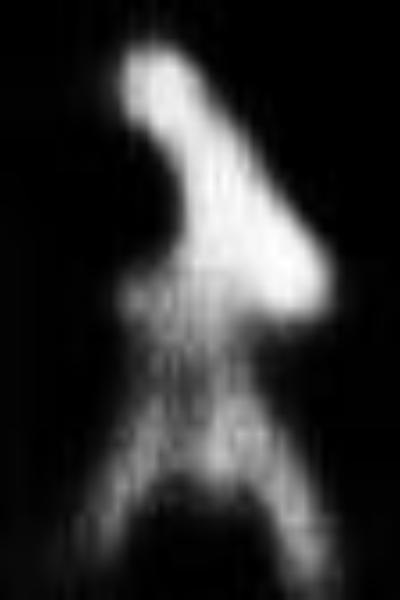} \\\vspace{1.5pt}
\includegraphics[width=1\columnwidth]{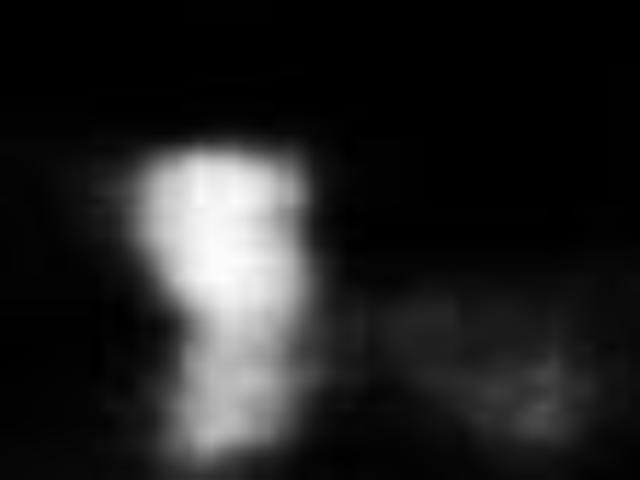} \\\vspace{1.5pt}
\end{minipage} 
}
\hspace{-12pt}
\subfigure[MPCI]{
\centering
\begin{minipage}{0.1\linewidth}
\centering
\includegraphics[width=1\columnwidth]{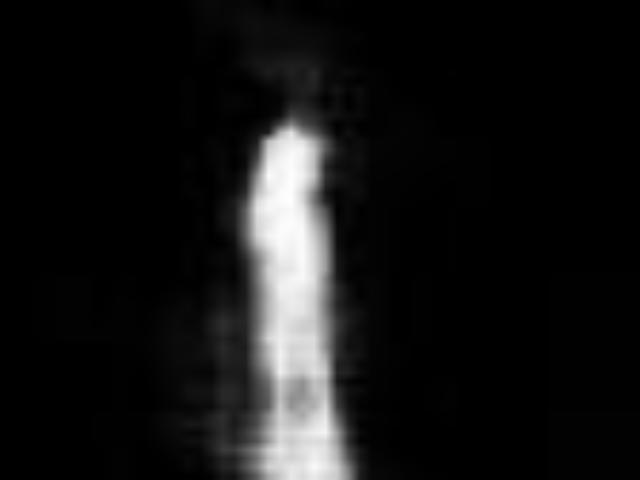} \\\vspace{1.5pt}
\includegraphics[width=1\columnwidth]{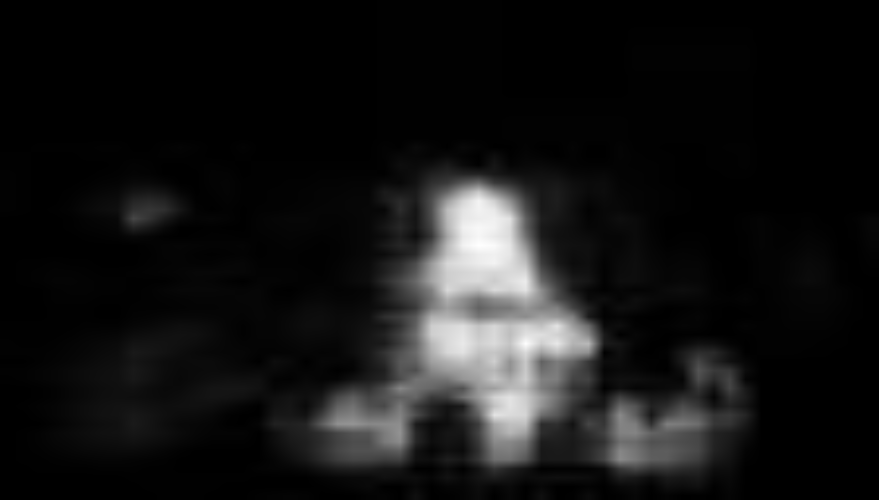} \\\vspace{1.5pt}
\includegraphics[width=1\columnwidth]{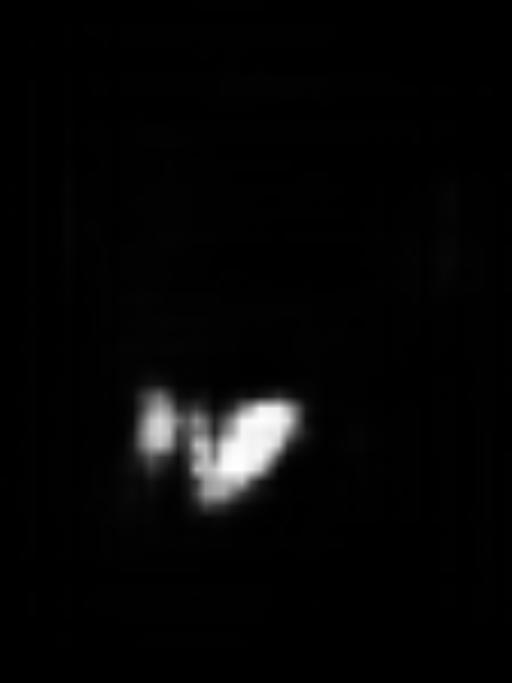} \\\vspace{1.5pt}
\includegraphics[width=1\columnwidth]{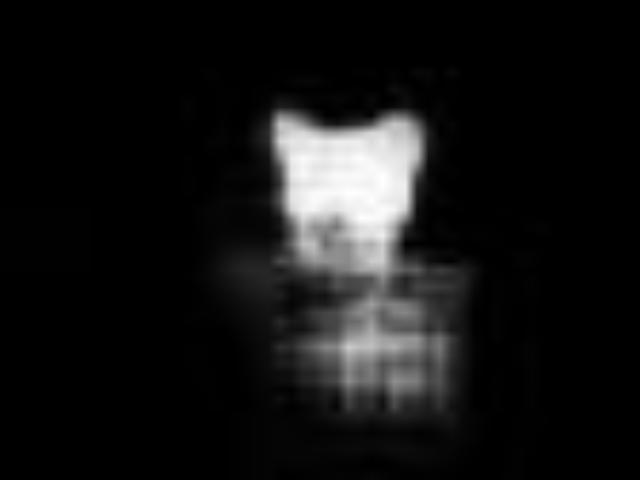} \\\vspace{1.5pt}
\includegraphics[width=1\columnwidth]{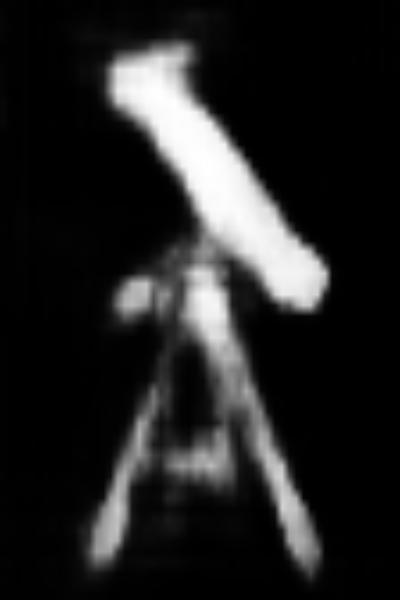} \\\vspace{1.5pt}
\includegraphics[width=1\columnwidth]{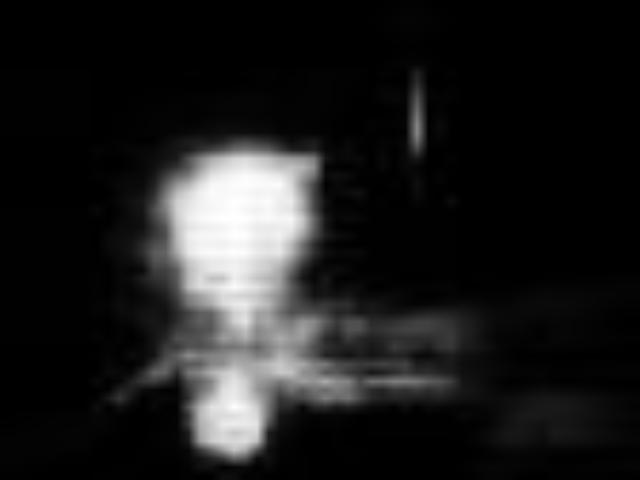} \\\vspace{1.5pt}
\end{minipage} 
}
\hspace{-12pt}
\subfigure[PCA]{
\centering
\begin{minipage}{0.1\linewidth}
\centering
\includegraphics[width=1\columnwidth]{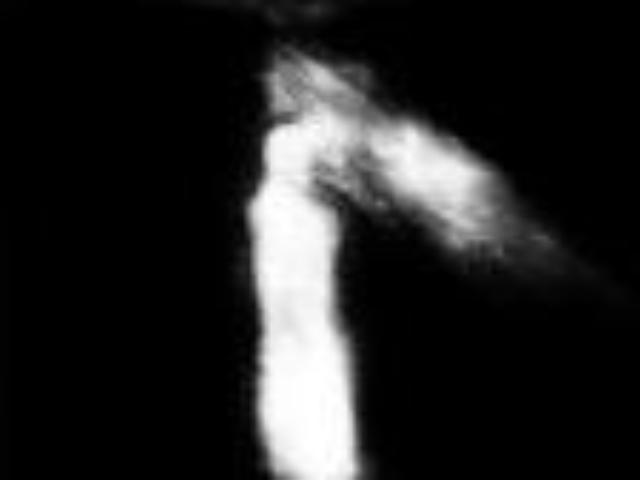} \\\vspace{1.5pt}
\includegraphics[width=1\columnwidth]{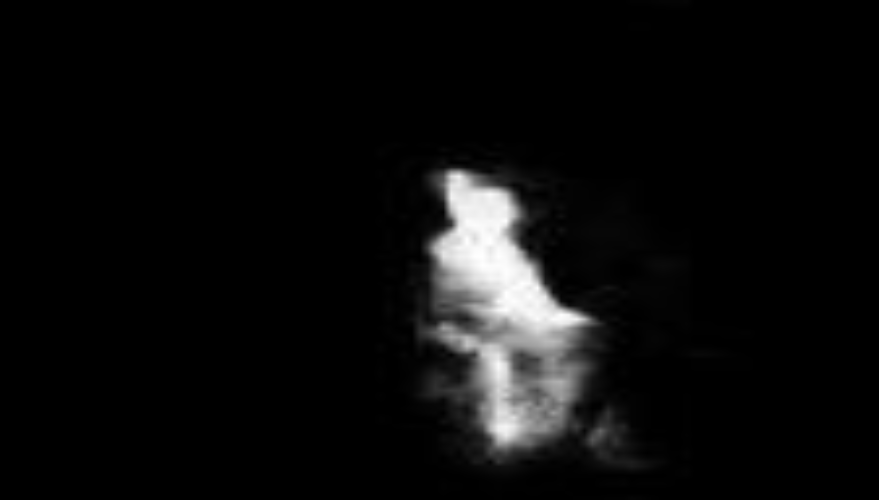} \\\vspace{1.5pt}
\includegraphics[width=1\columnwidth]{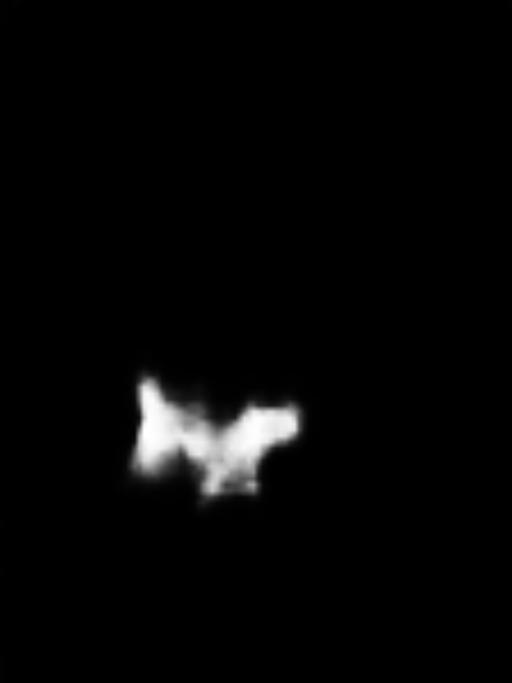} \\\vspace{1.5pt}
\includegraphics[width=1\columnwidth]{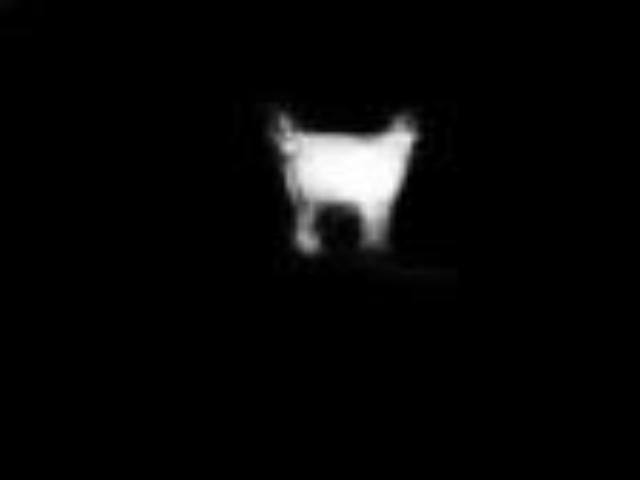} \\\vspace{1.5pt}
\includegraphics[width=1\columnwidth]{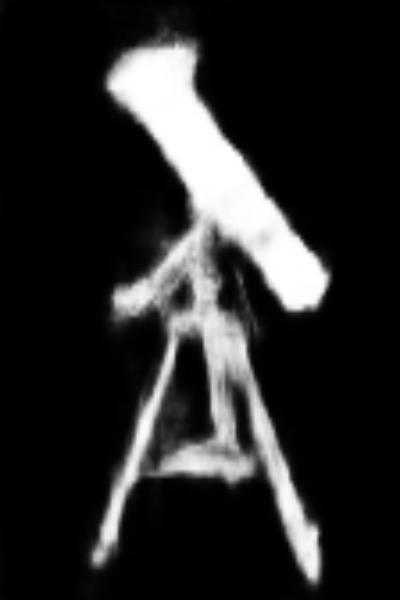} \\\vspace{1.5pt}
\includegraphics[width=1\columnwidth]{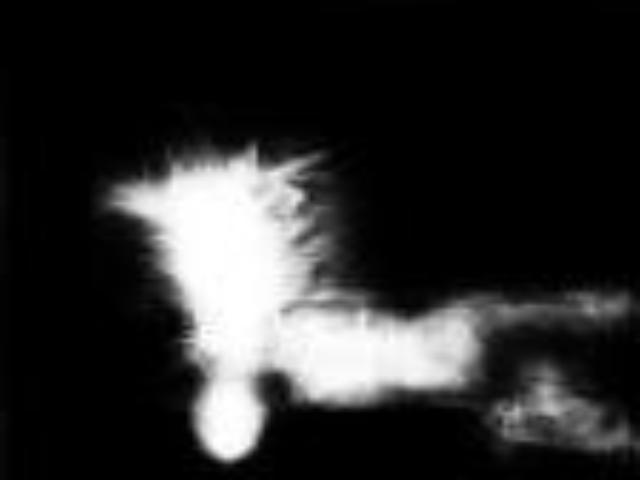} \\\vspace{1.5pt}
\end{minipage} 
}
\hspace{-12pt}
\subfigure[OURS]{
\centering
\begin{minipage}{0.1\linewidth}
\centering
\includegraphics[width=1\columnwidth]{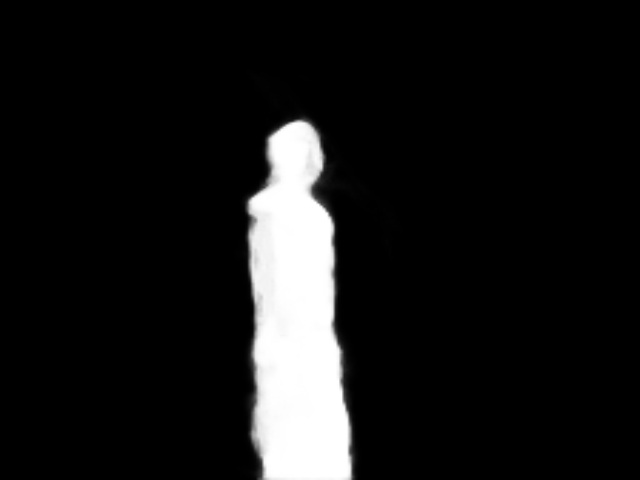} \\\vspace{1.5pt}
\includegraphics[width=1\columnwidth]{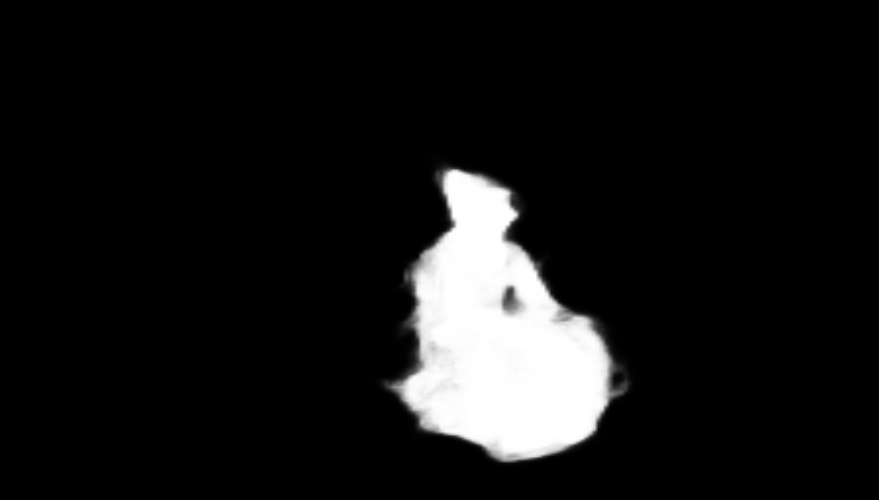} \\\vspace{1.5pt}
\includegraphics[width=1\columnwidth]{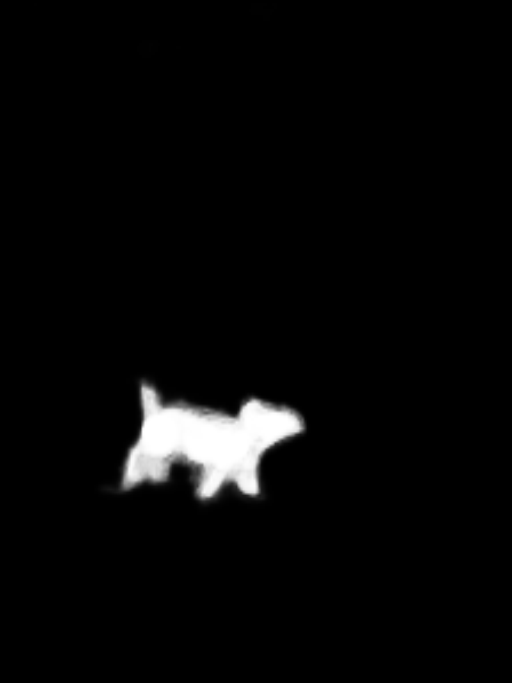} \\\vspace{1.5pt}
\includegraphics[width=1\columnwidth]{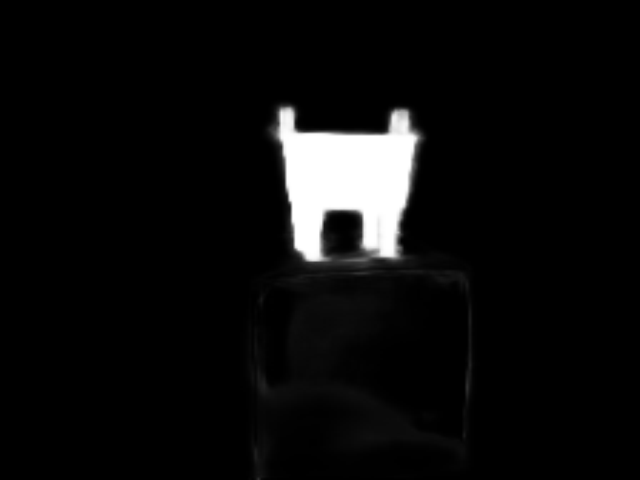} \\\vspace{1.5pt}
\includegraphics[width=1\columnwidth]{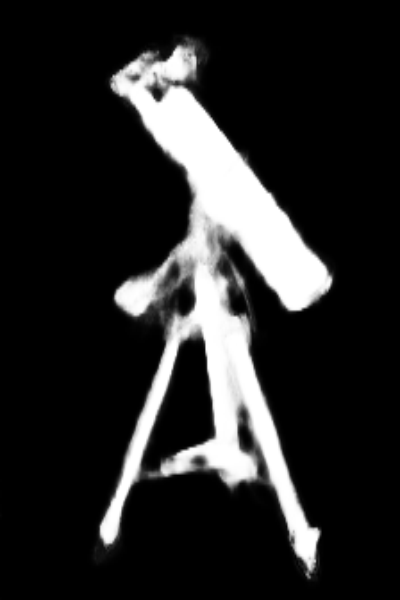} \\\vspace{1.5pt}
\includegraphics[width=1\columnwidth]{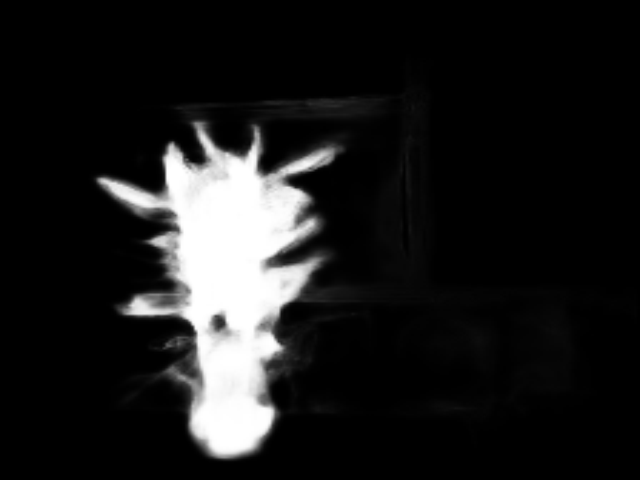} \\\vspace{1.5pt}
\end{minipage} 
}
\hspace{-12pt}
\subfigure[GT]{
\centering
\begin{minipage}{0.1\linewidth}
\centering
\includegraphics[width=1\columnwidth]{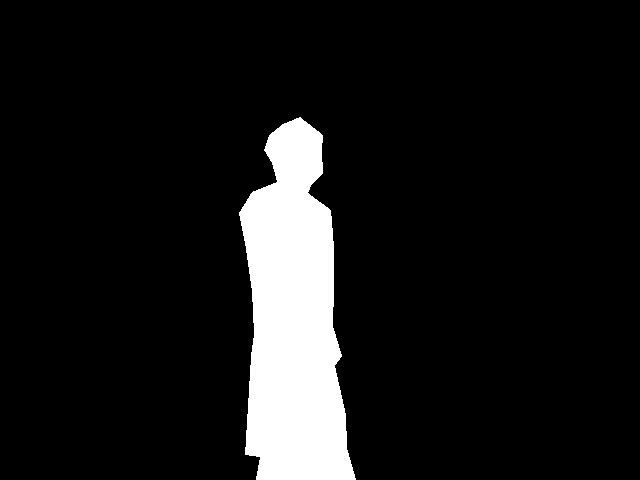} \\\vspace{1.5pt}
\includegraphics[width=1\columnwidth]{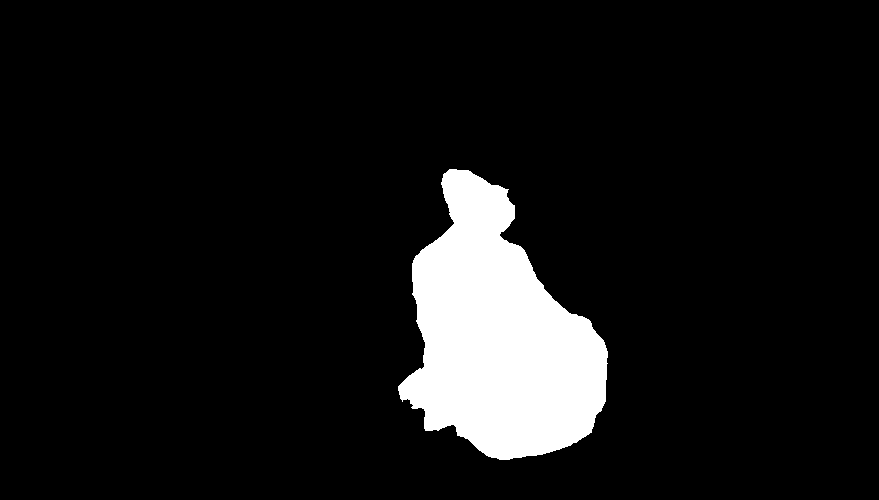} \\\vspace{1.5pt}
\includegraphics[width=1\columnwidth]{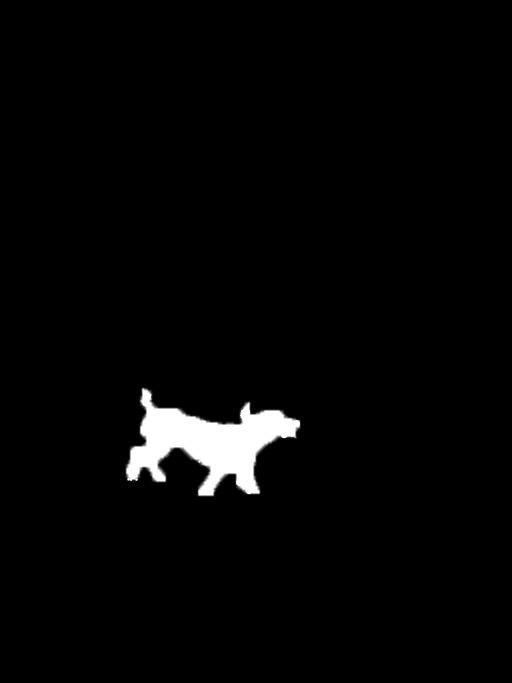} \\\vspace{1.5pt}
\includegraphics[width=1\columnwidth]{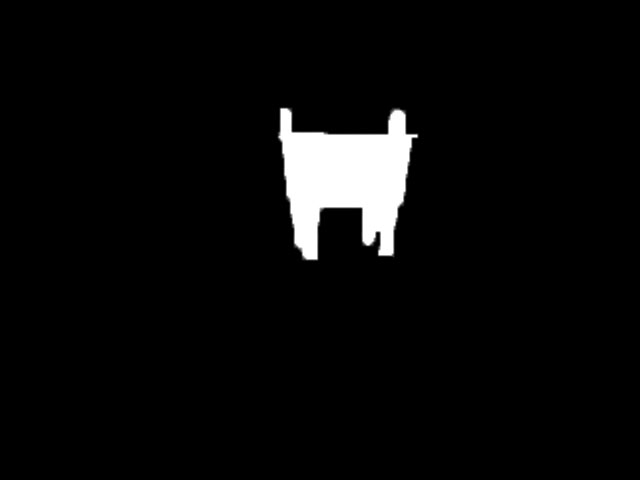} \\\vspace{1.5pt}
\includegraphics[width=1\columnwidth]{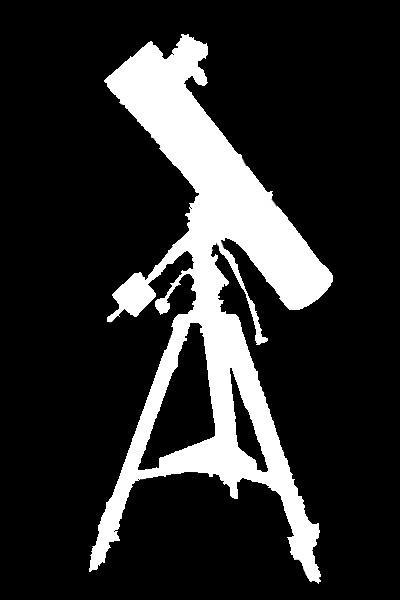} \\\vspace{1.5pt}
\includegraphics[width=1\columnwidth]{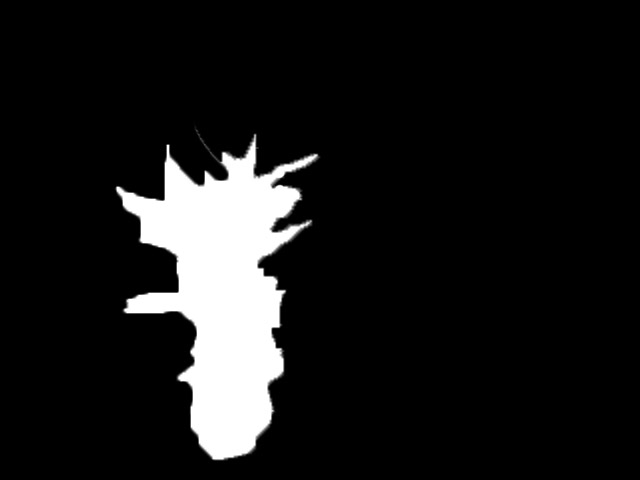} \\\vspace{1.5pt}
\end{minipage} 
}
\caption{Visual comparison of saliency maps.}     
\label{fig:results}     
\end{figure*}

Fig.~\ref{fig:results} provides a visual comparison between our model and other approaches. For the typical scenarios that share the similar appearance with the background, as shown in the first two rows, the proposed method can better capture effective information in depth data and localize the salient objects accurately. The depth distributions in the third and fourth rows are indistinguishable for the salient objects. Other methods fail to highlight complete and uniform salient objects while our fusion strategy can avoid such depth confusions to a great extent. Moreover, benefited from the edge-preserving loss, the proposed method preserves rich details and sharp boundaries in comparison with the others as demonstrated in the last two rows.

\subsubsection*{Failed Cases}
\begin{figure*}[htbp] 
\centering
\subfigure[RGB]{
\centering
\begin{minipage}{0.1\linewidth}
\centering
\includegraphics[width=1\columnwidth]{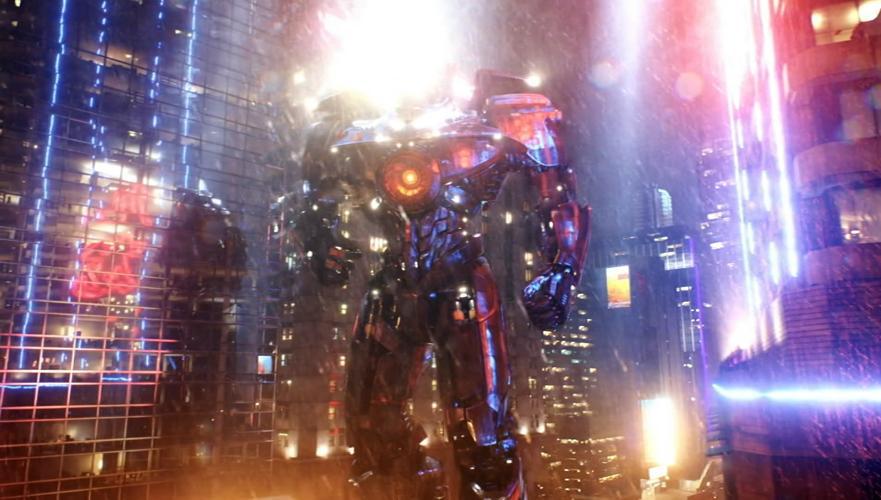} \\\vspace{1.5pt}
\includegraphics[width=1\columnwidth]{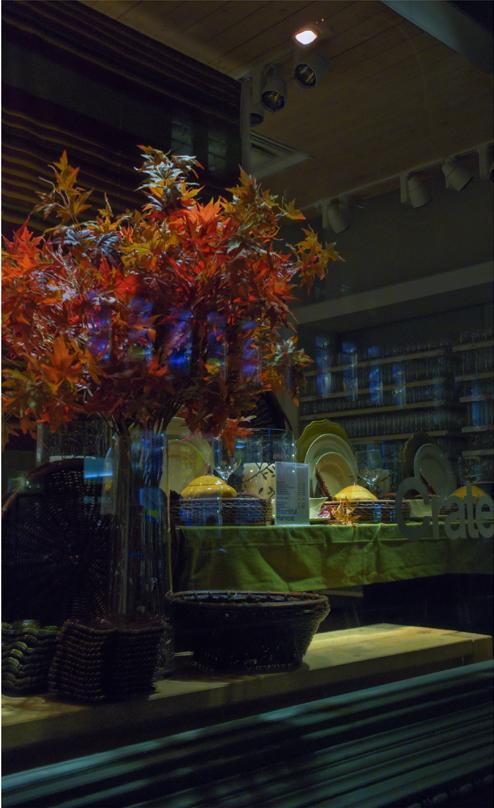} \\\vspace{1.5pt}
\end{minipage} 
}
\hspace{-12pt}
\subfigure[Depth]{
\centering
\begin{minipage}{0.1\linewidth}
\centering
\includegraphics[width=1\columnwidth]{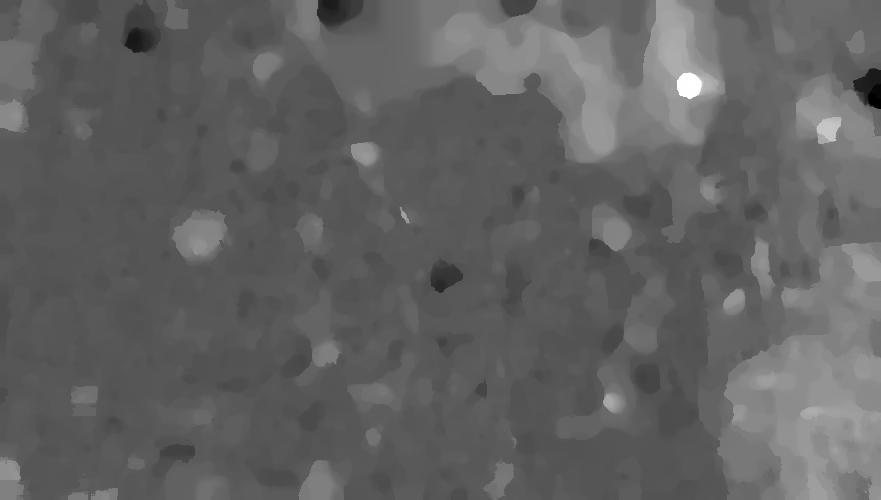} \\\vspace{1.5pt}
\includegraphics[width=1\columnwidth]{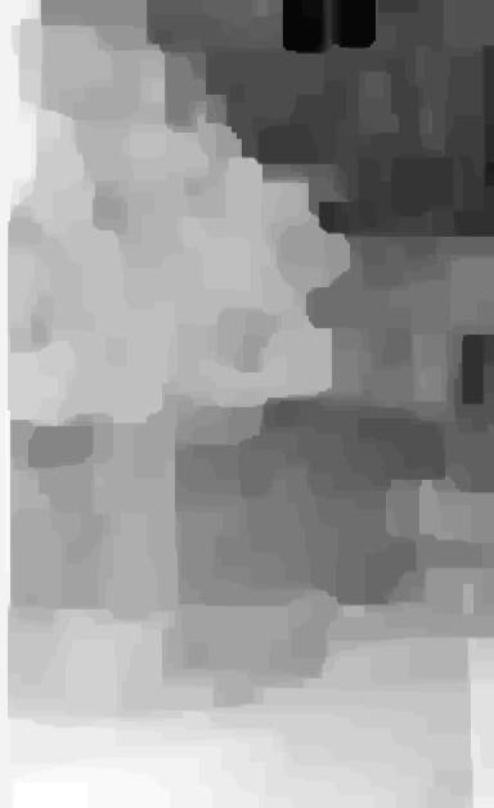} \\\vspace{1.5pt}
\end{minipage} 
}
\hspace{-12pt}
\subfigure[GP]{
\centering
\begin{minipage}{0.1\linewidth}
\centering
\includegraphics[width=1\columnwidth]{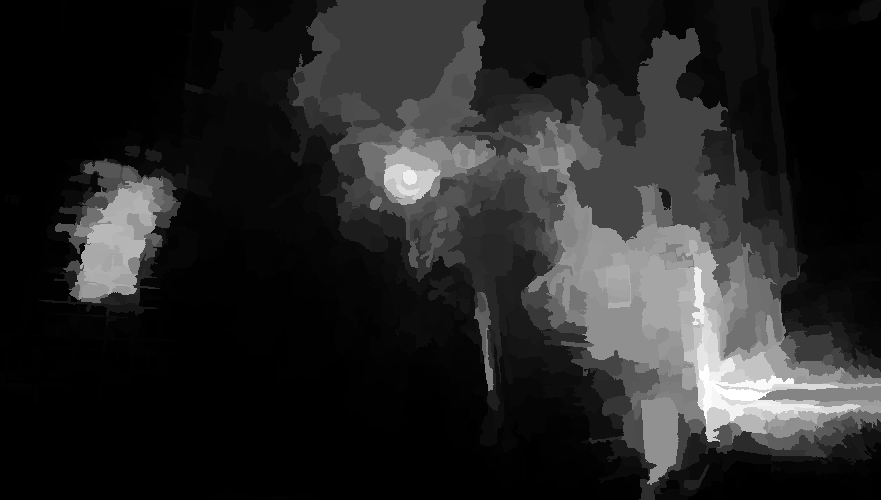} \\\vspace{1.5pt}
\includegraphics[width=1\columnwidth]{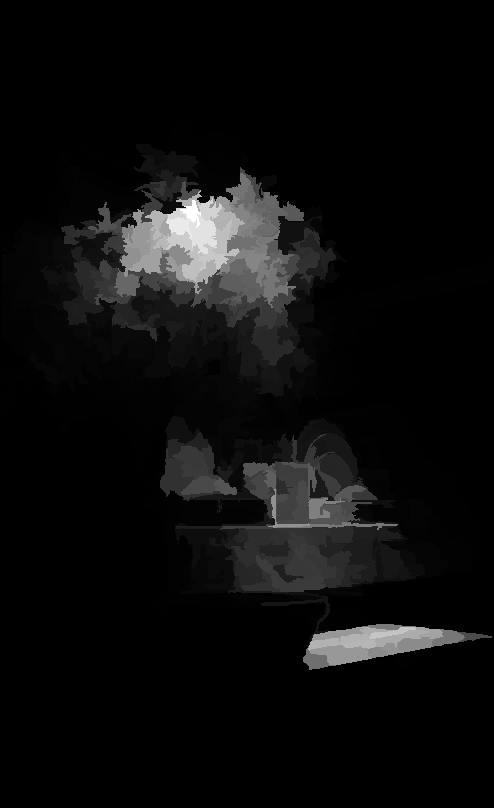} \\\vspace{1.5pt}
\end{minipage}
}
\hspace{-12pt}
\subfigure[LBE]{
\centering
\begin{minipage}{0.1\linewidth}
\centering
\includegraphics[width=1\columnwidth]{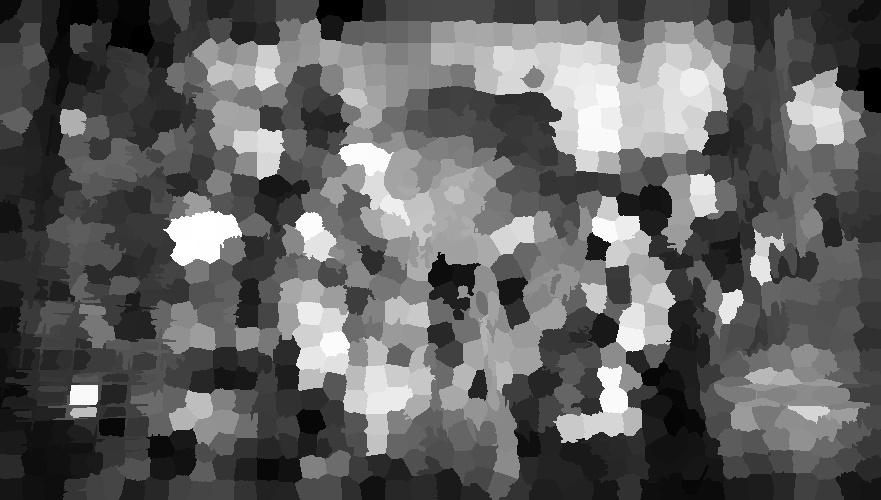} \\\vspace{1.5pt}
\includegraphics[width=1\columnwidth]{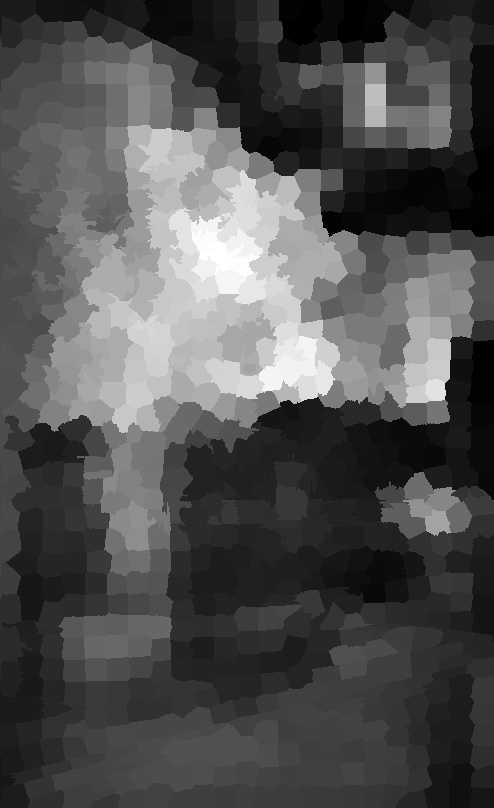} \\ \vspace{1.5pt}
\end{minipage}  
}
\hspace{-12pt}
\subfigure[CTMF]{
\centering
\begin{minipage}{0.1\linewidth}
\centering
\includegraphics[width=1\columnwidth]{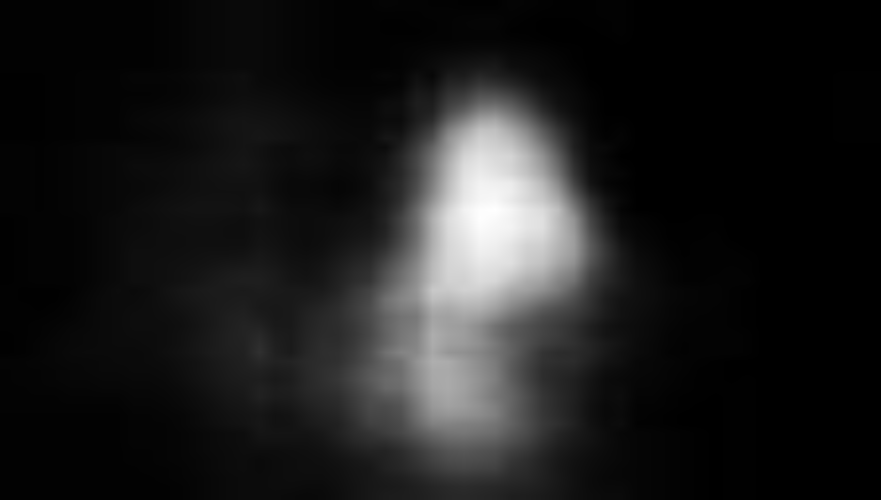} \\\vspace{1.5pt}
\includegraphics[width=1\columnwidth]{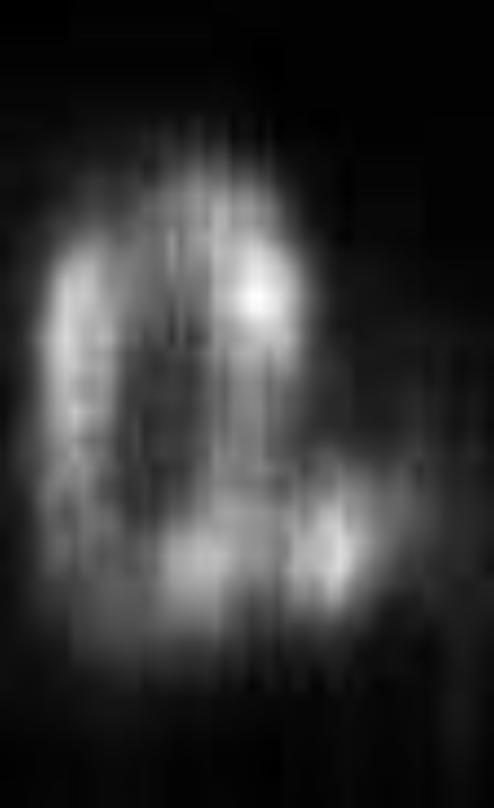} \\\vspace{1.5pt}
\end{minipage} 
}
\hspace{-12pt}
\subfigure[MPCI]{
\centering
\begin{minipage}{0.1\linewidth}
\centering
\includegraphics[width=1\columnwidth]{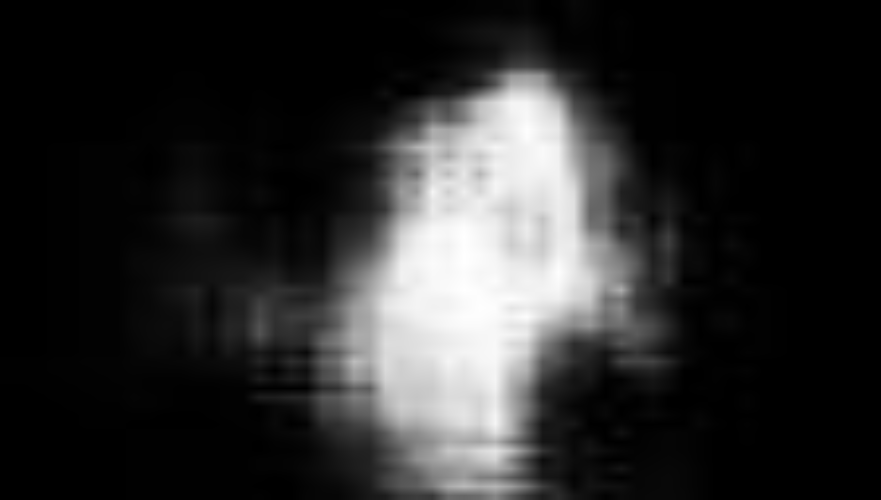} \\\vspace{1.5pt}
\includegraphics[width=1\columnwidth]{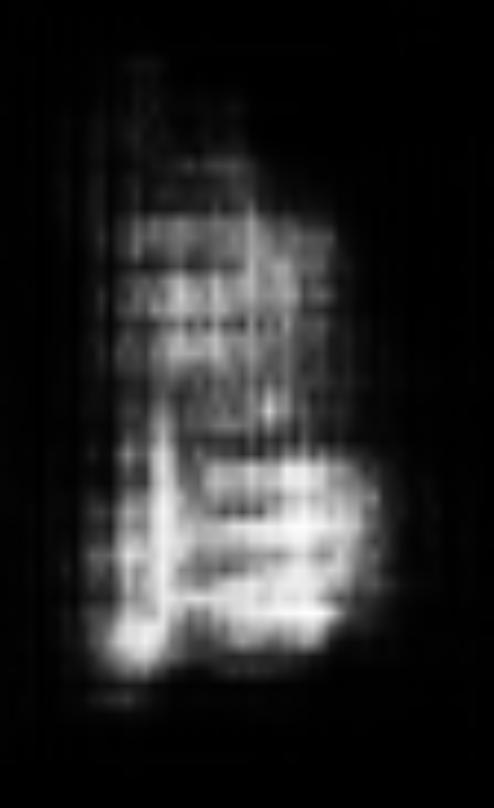} \\\vspace{1.5pt}
\end{minipage} 
}
\hspace{-12pt}
\subfigure[PCA]{
\centering
\begin{minipage}{0.1\linewidth}
\centering
\includegraphics[width=1\columnwidth]{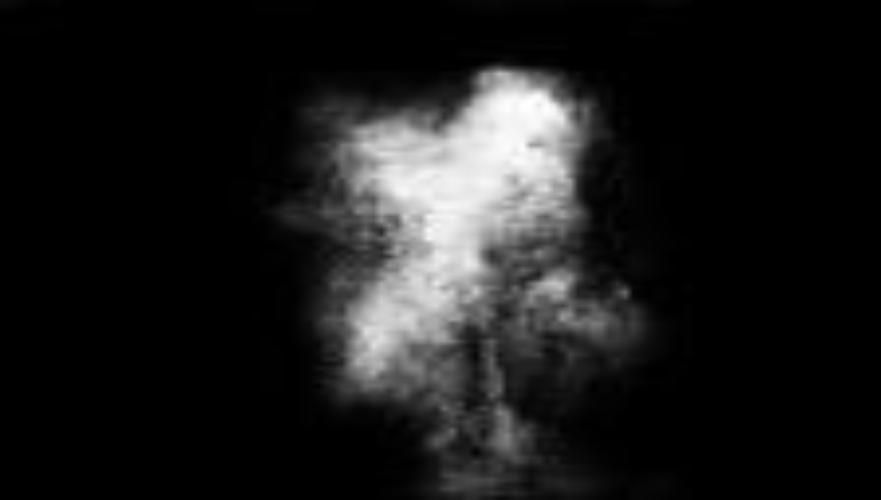} \\\vspace{1.5pt}
\includegraphics[width=1\columnwidth]{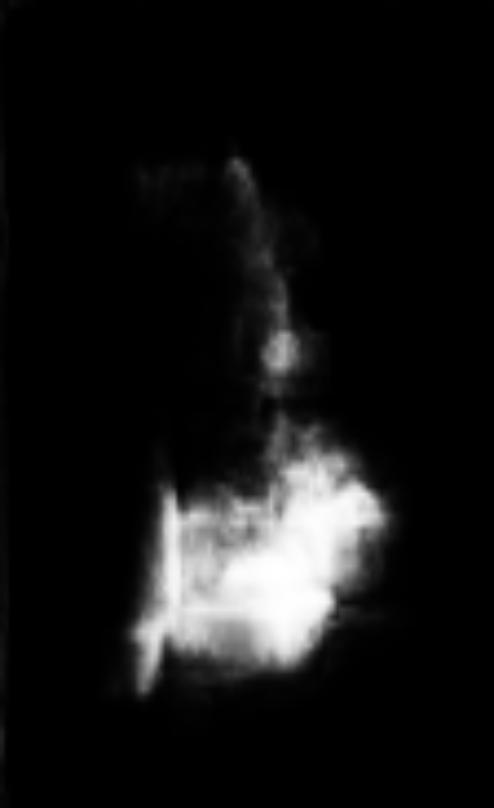} \\\vspace{1.5pt}
\end{minipage} 
}
\hspace{-12pt}
\subfigure[OURS]{
\centering
\begin{minipage}{0.1\linewidth}
\centering
\includegraphics[width=1\columnwidth]{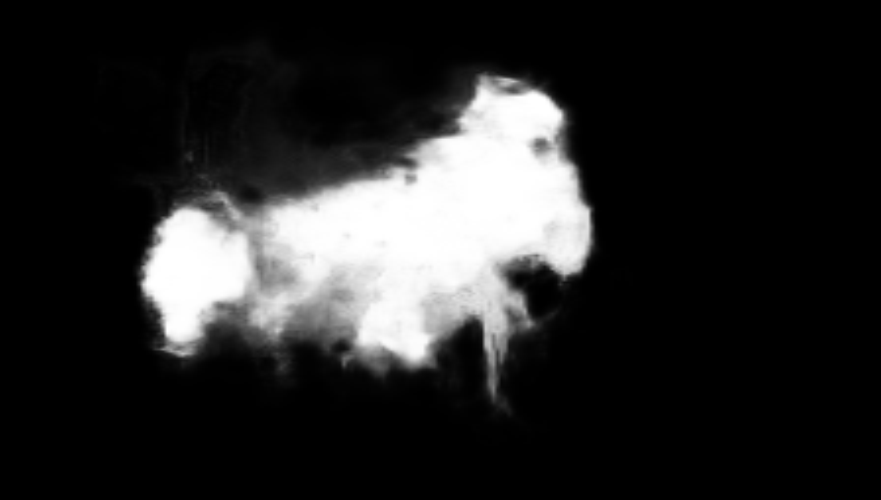} \\\vspace{1.5pt}
\includegraphics[width=1\columnwidth]{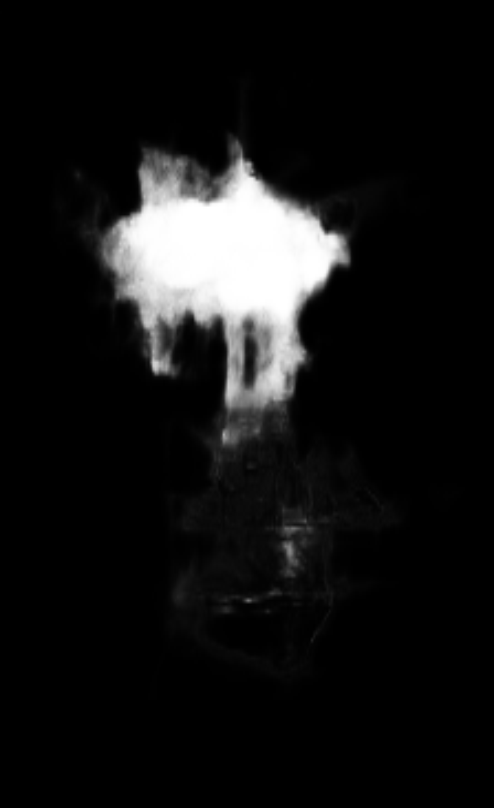} \\\vspace{1.5pt}
\end{minipage} 
}
\hspace{-12pt}
\subfigure[GT]{
\centering
\begin{minipage}{0.1\linewidth}
\centering
\includegraphics[width=1\columnwidth]{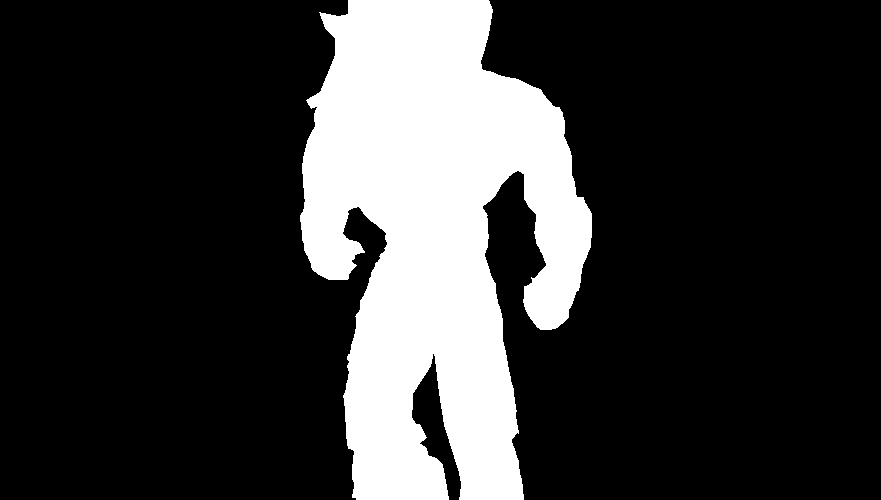} \\\vspace{1.5pt}
\includegraphics[width=1\columnwidth]{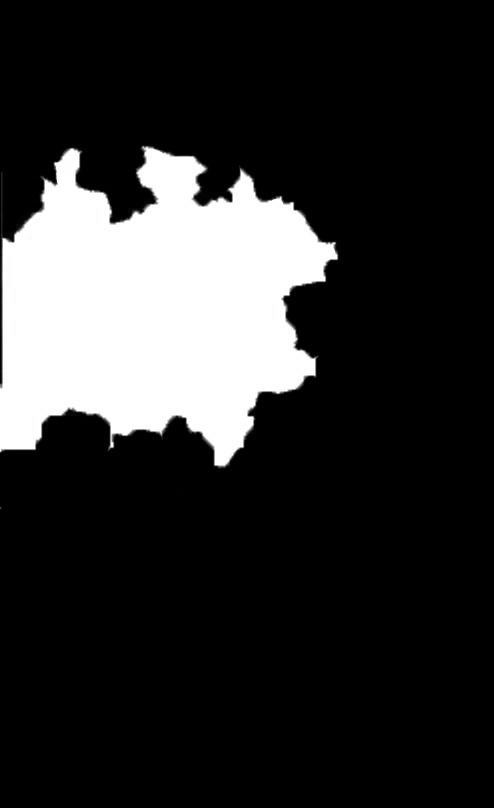} \\\vspace{1.5pt}
\end{minipage} 
}
\caption{Failed cases.}     
\label{fig:failedcases}     
\end{figure*}

The proposed approach is capable of detecting salient objects as long as the objects stand out in one modality. When objects are not distinguishable in both modalities, our approach fails as expected. Fig.~\ref{fig:failedcases} demonstrates two typical examples. As shown in the figure, such scenarios are challenging to all existing methods. 

\section{Conclusion}
In this paper, we have presented a novel end-to-end framework for RGB-D salient object detection. Instead of directly concatenating RGB and depth features or element-wisely multiplying/adding saliency predictions, we introduce a switch map that is adaptively learned to fuse the effective information from RGB and depth predictions. An edge-preserving loss is also designed for correcting blurry boundaries and further improving spatial coherence. The experiments have demonstrated that the proposed method consistently outperforms other state-of-the-art methods on different datasets.

\section*{Acknowledgement}
This work was supported in part by Major Scientific Project of Zhejiang Lab (No. 2018DD0ZX01) and the Natural Science Foundation of Zhejiang Province, China under Grant LY17F010007.

\bibliographystyle{IEEEtran}
\bibliography{IEEEabrv,./egbib}

\end{document}